\documentclass[10pt,twocolumn,letterpaper]{article}

\usepackage{wacv}              

\definecolor{wacvblue}{rgb}{0.21,0.49,0.74}
\usepackage[pagebackref,breaklinks,colorlinks,allcolors=wacvblue]{hyperref}
\usepackage{array}
\usepackage{multirow}
\usepackage{dblfloatfix}
\usepackage{placeins}
\def\wacvPaperID{185} 
\def\confName{WACV}
\def\confYear{2027}

\title{Continuous Angular Control in Text-to-Image Generation via Geometric Token Encoding}

\author{
Furkan G\"uzelant$^{1}$ \quad
Basavaraj Sunagad$^{2}$ \quad
Aysegul Dundar$^{1}$ \quad
Adam Kortylewski$^{3}$ \\
$^{1}$Bilkent University, Turkey \\
$^{2}$CISPA Helmholtz Center for Information Security, Germany \\
$^{3}$University of Freiburg, Germany \\[2pt]
{\footnotesize
 \texttt{furkan.guzelant@bilkent.edu.tr} \quad
\texttt{basavaraj.sunagad@cispa.de}} \\
{\footnotesize
\texttt{adundar@cs.bilkent.edu.tr} \quad
\texttt{kortylewski@cs.uni-freiburg.de}
}
}

\begin{document}
\twocolumn[{%
\renewcommand\twocolumn[1][]{#1}%
\maketitle
\vspace{-1cm}

\begin{center}
    \small\url{https://furkanguzelant.github.io/angular_control/}
\end{center}

\begin{center}
    \centering
    \captionsetup{type=figure}
    \includegraphics[width=0.99\textwidth]{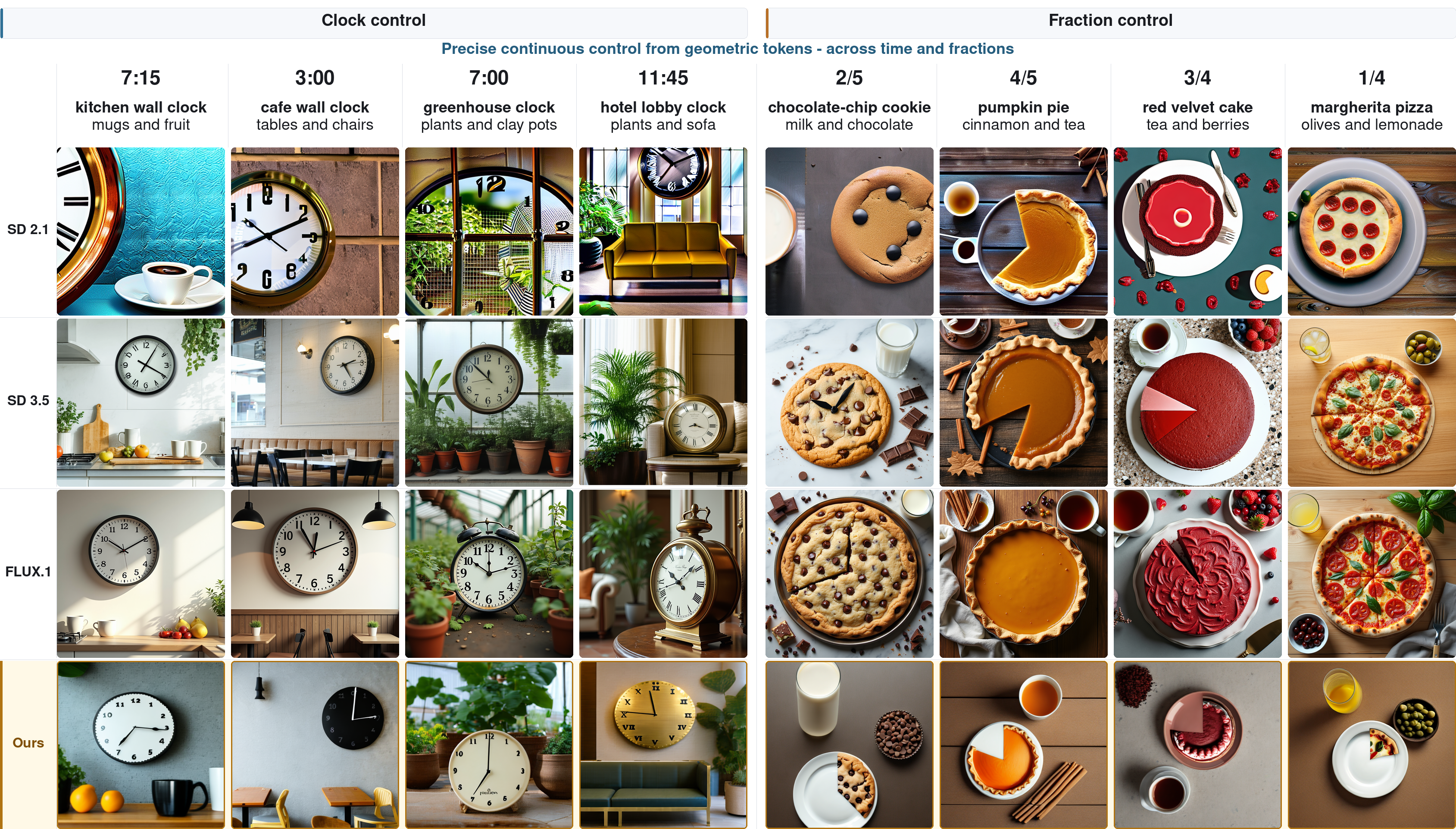}
    \captionof{figure}{
    \textbf{Continuous geometric control for clocks and fractions.}
    Existing text-to-image models often fail to follow the requested geometric quantity, producing common or canonical configurations instead.
    Our method accurately controls analog clock time and circular fractions across diverse scenes and object appearances. Prompts are abbreviated for readability; full generation prompts are provided in Appendix E of the supplementary material.
    }
    \label{fig:teaser}
\end{center}%
}]

\thispagestyle{plain}
\pagestyle{plain}

\begin{abstract}

Text-to-image diffusion models can render a photorealistic analog clock but
cannot control the time it displays. Across four pretrained models, fewer than
$1\%$ of generated clocks place both hands within $6^\circ$ of the requested
time; the models instead collapse onto visually dominant configurations such
as the canonical 10:10 pose. The same failure appears when a specific fraction
of a circular object is requested, where models return complete or
conventionally sliced objects. We observe that these two failures share a
geometric structure: both target quantities can be parameterized by angular variables.

We introduce a geometric token encoding framework that conditions generation
directly on these angles. Each angular component is encoded as a periodic
sine--cosine feature and projected into a role-specific token embedding that is injected into the prompt before a frozen text encoder. To spatially ground these tokens without requiring localization input at inference, we introduce Stochastic Localization Relaxation (SLR), which stochastically relaxes attention supervision from component-level to object-level to unconstrained localization during training. For clocks, we additionally introduce a hand-swap preference objective that discourages assigning the correct angles to the wrong hand roles. On analog clocks our method reaches $63$--$78\%$ Acc@$6^\circ$ across evaluation methods, compared with below $1\%$ for the baselines. On circular-fraction generation, it achieves $89.5\%$ Acc@$0.10$, while maintaining $88.8\%$ on unseen object categories and $84.9\%$ on unseen target values, compared with roughly $20\%$ for pretrained baselines.

\end{abstract}
 
\section{Introduction}
\label{sec:introduction}

Strong text-to-image models can generate photorealistic analog clocks, but
cannot reliably control the time they display. Across Stable Diffusion~2.1,
SDXL, Stable Diffusion~3.5, and Flux.1~\cite{rombach2022highresolution,podell2024sdxl,
esser2024scaling,flux2024}, fewer than $1\%$ of the generated clocks place
both hands within $6^\circ$ of the requested time. As illustrated in Figure~\ref{fig:teaser}, the models
frequently produce common clock configurations, with a pronounced bias toward
canonical poses such as 10:10, which we quantify in Appendix~C.
Similarly, when generating a requested fraction
of a circular object, models often produce a complete object or a conventional
slice rather than the specified amount. In both cases, the image is plausible and shows the correct object category but not the requested quantitative value.

These observations expose a limitation of natural-language conditioning.
Language is effective for describing semantic content, appearance, and scene
context, but provides an unreliable interface for precise numerical and
geometric control. A model may understand that the requested object is a
clock or a pizza without faithfully representing the particular time or
fraction specified in the prompt.
At first glance, clock time and circular fractions appear to be unrelated
generation problems. We observe, however, that they share the same underlying
geometric structure: the requested quantity is represented by one or more
angles on a circle. Clock time is jointly determined by the orientations of
the hour and minute hands, while a circular fraction is specified by the
angular boundaries of its visible sector. This common structure suggests that
these quantities should be represented directly in angular space rather than
expressed only through language.

Existing controllable-generation methods either require explicit structural
inputs or encode continuous camera and object-pose
parameters~\cite{zhang2023adding,cheng2024continuous,
rishubh2025compasscontrol}. A structural input depicting the requested clock
hands or circular sector would already specify the desired output, while
existing parametric methods primarily control extrinsic properties of complete
objects. We instead generate an intrinsic quantitative state directly from its
numerical value by coordinating multiple geometric components within the same
visual structure.
We introduce a geometric token framework that conditions text-to-image
generation directly on continuous angular quantities. Target angles are
encoded as periodic sine-cosine features and projected into role-specific
tokens, letting the model represent precise geometry while preserving each
component's semantic function.

To learn spatial associations without requiring localization input at test
time, we build on the attention-localization strategy~\cite{rishubh2025compasscontrol} and introduce
\emph{Stochastic Localization Relaxation (SLR)}. By varying the spatial supervision
available during training, this strategy prevents the model from becoming
dependent on object-specific bounding boxes. At inference, generation requires
only the text prompt and target quantity. For clock generation, we additionally
introduce a hand-swap preference objective that discourages assigning the
correct angles to the wrong hand roles.

Across both tasks, our method enables reliable control of quantitative visual
states that pretrained text-to-image models rarely generate correctly from
language alone. It raises clock-time accuracy from under $1\%$ to
$63$--$78\%$ and fraction accuracy from roughly $20\%$ to $89.5\%$, while
generalizing to unseen visual appearances, object categories, and continuous
target values.
Our main contributions are:

\begin{itemize}

    \item We identify the shared angular structure of clock-time and
    circular-fraction generation, and introduce a periodic, role-specific
    token representation for the coordinated components of an intrinsic
    visual state.

    \item We introduce Stochastic Localization Relaxation (SLR), which uses localized
    attention supervision during training while enabling bounding-box-free
    inference. For clock generation, we further introduce a hand-swap
    preference objective that improves semantic-role assignment between the
    hour and minute hands.

    \item We evaluate on analog clocks and circular fractions, improving
    Acc@$6^\circ$ from below $1\%$ to $63$--$78\%$ over pretrained
    baselines, and generalize to unseen appearances and continuous target
    values.

\end{itemize}

\section{Related Work}
\label{sec:related_work}

\noindent\textbf{Controllability in text-to-image diffusion.}
Text-to-image diffusion models achieve strong photorealism and text-image
alignment by conditioning the denoising process on natural-language
representations~\cite{rombach2022highresolution,podell2024sdxl}.
However, language provides only an indirect interface for specifying exact
numerical and geometric configurations. Requests involving angles, clock
times, or fractions may therefore be ignored or replaced by visually common
configurations. This difficulty is well documented for discrete quantities:
generating an exact number of objects requires dedicated counting
mechanisms~\cite{binyamin2025make,kang2025counting}, and compositional
benchmarks report persistent failures~\cite{huang2023t2icompbench,huang2025t2icompbenchpp}.
We study the continuous counterpart, where the requested quantity is a
geometric configuration rather than a count.

A broad line of work improves controllability by manipulating text embeddings,
model parameters, or attention maps~\cite{gal2022textual,gandikota2024concept,baumann2025continuous,
chiu2026textslider,ye2026allinone,ge2023richtext,epstein2023selfguidance,
bartal2023multidiffusion,cao2023masactrl}.
Another approach provides explicit structural conditions.
ControlNet~\cite{zhang2023adding} and
T2I-Adapter~\cite{mou2024t2i} condition generation on inputs such as edges,
depth, segmentation, and pose, while grounded generation methods use bounding
boxes or attention constraints to control object
placement~\cite{li2023gligen,xie2023boxdiff}.
Although effective, these methods require either learned semantic controls or
additional structural inputs. In our setting, an edge or pose map of the
requested clock hands would already specify the target. We instead generate
the desired geometry directly from a compact numerical input.

\noindent\textbf{Continuous geometric conditioning.}
Learned-token and textual-inversion methods~\cite{gal2022textual,voynov2023pplus,alaluf2023neti,
zhang2023prospect} demonstrate that visual concepts can be represented
directly in the text-conditioning space. Subsequent methods use
this interface to represent continuously varying geometric or semantic
parameters. ViewNeTI~\cite{burgess2025viewpoint} learns embeddings for
continuous camera viewpoints, while Continuous 3D
Words~\cite{cheng2024continuous} controls 3D-aware attributes such as object
orientation and illumination. Camera Control~\cite{lu2026viewtoken} similarly
maps continuous camera parameters to viewpoint tokens.
Related work also explores continuous semantic control through learned
directions and lightweight adapters~\cite{gandikota2024concept,
chiu2026textslider,ye2026allinone,baumann2025continuous}.
These methods mainly modulate semantic appearance attributes rather than
metrically defined quantities represented by multiple coordinated geometric
components.

\noindent\textbf{Object orientation and spatial grounding.}
Compass Control~\cite{rishubh2025compasscontrol} is the closest architectural
basis for our work. It maps continuous object orientations to learned
geometric tokens and introduces Coupled Attention Localization (CALL) to
associate each token with its corresponding object region. This enables
independent orientation control for multiple objects in a scene.
ORIGEN~\cite{min2025origen} considers the related problem of object-orientation
grounding through zero-shot guidance from a pretrained orientation estimator,
while LooseControl~\cite{bhat2024loosecontrol} uses loose depth conditioning
for 3D-aware scene control.
These methods control the orientation or placement of complete objects. We
instead consider intrinsic semantic quantities represented by coordinated
components within a single visual structure: clock time is jointly determined
by the hour and minute hands, and a circular fraction by its angular
boundaries.

Text-to-image models are also known to exhibit semantic and attribute-binding
failures~\cite{chefer2023attend,feng2023trainingfree}. In our setting, this
takes a geometric form: numerical values must remain associated with their
correct structural roles. A clock may contain both requested angles but still
be wrong if the hour- and minute-hand values are exchanged. Our hand-swap
objective is related to preference-based diffusion fine-tuning such as
Diffusion-DPO~\cite{wallace2024diffusiondpo}, but uses swapped geometric
conditionings as hard negatives. Our method further combines periodic
role-specific tokens with training-time localization, without requiring
bounding boxes at inference.
\section{Method}
\label{sec:method}

Our framework controls quantities defined by coordinated
geometric components. Given a prompt and target quantity,
we encode its angles periodically and project them
into role-specific geometric tokens.
These tokens replace placeholder embeddings at semantic
positions in the prompt before the frozen text encoder,
whose output conditions a LoRA-adapted Stable Diffusion
2.1 (SD 2.1) backbone.

\begin{figure*}[t]
  \centering
  \begin{minipage}[t]{0.60\linewidth}
    \includegraphics[width=\linewidth]{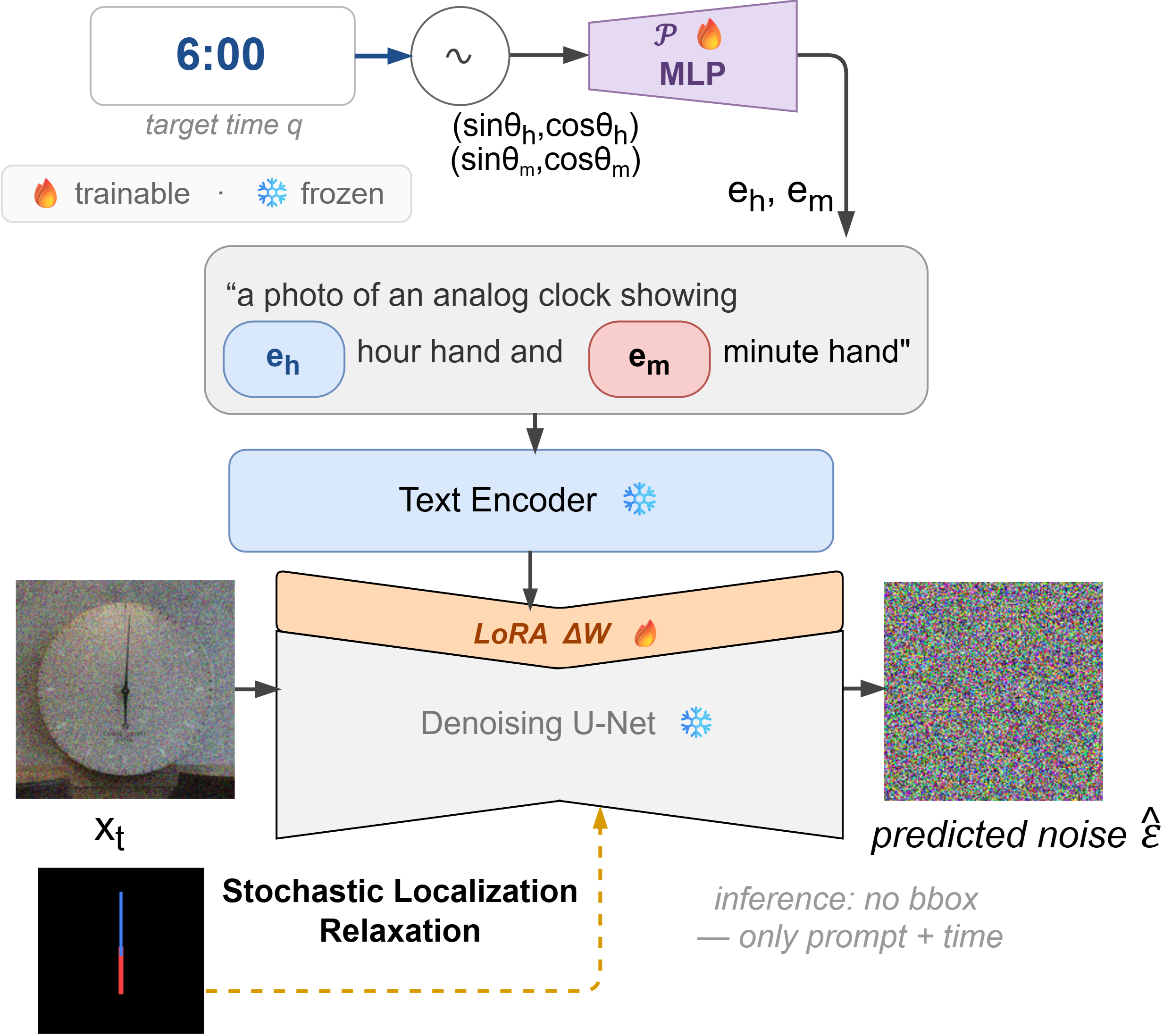}
  \end{minipage}
  \hfill
  \begin{minipage}[t]{0.39\linewidth}
    \includegraphics[width=\linewidth]{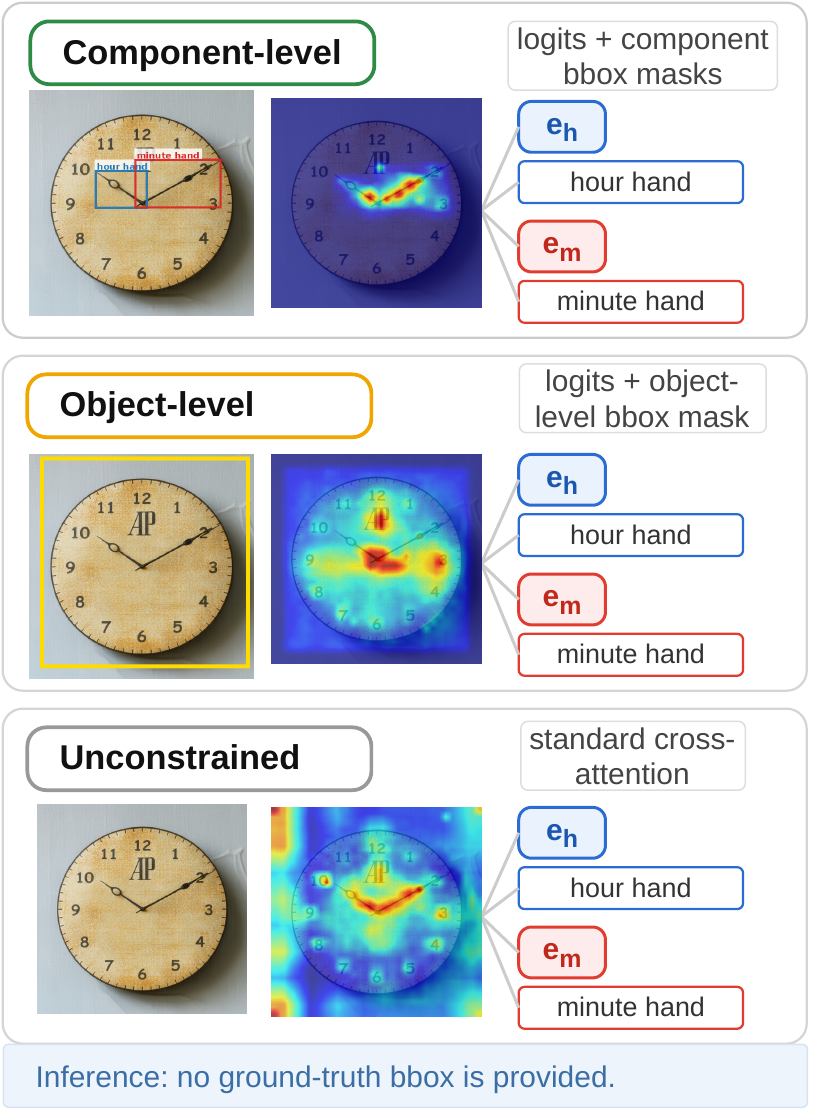}
  \end{minipage}
  \caption{
    \textbf{Left:} Overview of our geometric token encoding framework, illustrated for clock-time control.
    The target clock time $(h,m)$ or fraction-sector boundaries are encoded as sinusoidal pairs and mapped through the projection network~$P$ to produce role-specific geometric tokens $\mathbf{e}_h, \mathbf{e}_m$ (or $\mathbf{e}_s, \mathbf{e}_e$) injected into the prompt before the frozen text encoder.
    Both tasks share the same MLP architecture and condition a LoRA-adapted SD 2.1 denoising U-Net.
    \textbf{Right:} Stochastic Localization Relaxation (SLR).
    Training samples one of three localization regimes (55/30/15\%): component-level CALL using component boxes, object-level CALL using a shared object box, and unconstrained standard cross-attention. No localization input is required at inference.
  }
  \label{fig:method}
\end{figure*}

A central challenge is to ensure that each geometric token affects the correct
visual component without requiring spatial annotations at inference. We address
this using Stochastic Localization Relaxation (SLR), which provides stochastic spatial supervision during training, ranging from component-level
localization to fully unconstrained attention. At test time, generation uses
only the text prompt and target quantity. For clock generation, we additionally introduce a hand-swap preference
objective that reduces confusion between the hour- and minute-hand roles. Figure~\ref{fig:method} summarizes the complete
pipeline.

\subsection{Periodic Encoding and Role-Specific Geometric Tokens}
\label{sec:geometric_conditioning}

To condition the diffusion model on a requested geometric quantity, we first
convert the numerical target into a continuous representation that respects
its angular structure, and then map this representation into token embeddings
compatible with the pretrained text-to-image model.

\noindent\textbf{Periodic encoding.}
In both applications, the target quantity decomposes into angular components:
a clock time into the orientations of its hour and minute hands, a circular
fraction into the directions of its sector boundaries. Both tasks therefore
share the same geometric encoding.

Representing an angle as a scalar is problematic because angles do not lie on
a linear domain: values close to $0$ and $2\pi$ denote nearly identical
directions, yet their scalar values are far apart. Discretizing angles into
classes instead introduces quantization boundaries and prevents smooth control
between neighboring values.

We therefore encode each angular component
$\theta \in [0,2\pi)$ as

\begin{equation}
    \phi(\theta)
    =
    \begin{bmatrix}
        \sin\theta \\
        \cos\theta
    \end{bmatrix}
    \in \mathbb{R}^{2}.
    \label{eq:periodic_encoding}
\end{equation}

This representation is continuous across the circular boundary and maps
nearby directions to nearby points in the encoding space. For a target
containing $N$ angular components $\{\theta_i\}_{i=1}^{N}$, we concatenate
their representations:

\begin{equation}
    \mathbf{s}
    =
    \left[
        \phi(\theta_1)
        \,\Vert\,
        \phi(\theta_2)
        \,\Vert\,
        \cdots
        \,\Vert\,
        \phi(\theta_N)
    \right]
    \in \mathbb{R}^{2N},
    \label{eq:multi_angle_encoding}
\end{equation}

where $\Vert$ denotes concatenation.

\noindent\textbf{Role-specific geometric tokens.}
Although $\mathbf{s}$ provides a continuous description of the target
geometry, it cannot be supplied directly to the pretrained text-to-image
model, whose conditioning interface expects text-token embeddings.
Furthermore, the model must distinguish the semantic roles of the controlled
components. For example, generating the correct two clock-hand angles is
insufficient if their hour- and minute-hand assignments are exchanged.

We therefore use a lightweight projection network $P$ to map the joint
periodic representation to $N$ role-specific geometric tokens:

\begin{equation}
    P(\mathbf{s})
    =
    \left[
        \mathbf{e}_1,
        \mathbf{e}_2,
        \ldots,
        \mathbf{e}_N
    \right],
    \qquad
    \mathbf{e}_i \in \mathbb{R}^{d},
    \label{eq:geometric_projection}
\end{equation}

where $d$ is the dimensionality of the text-token embedding space. Because
$P$ receives the complete geometric representation, each output token can
represent its assigned component in the context of the other components that
jointly define the target.

\noindent\textbf{Task parameterization.}
For a clock time $(h,m)$, with angles measured clockwise from 12 o'clock, we
define

\begin{equation}
    \theta_h
    =
    2\pi\frac{(h\bmod 12)+m/60}{12},
    \qquad
    \theta_m
    =
    2\pi\frac{m}{60}.
    \label{eq:clock_angles}
\end{equation}

The resulting periodic representation is projected into an hour-hand token
$\mathbf{e}_h$ and a minute-hand token $\mathbf{e}_m$. For a circular
fraction $f$, we use a fixed start angle $\theta_s=0$ for all samples and an
end angle $\theta_e=2\pi f$, producing a start-boundary token
$\mathbf{e}_s$ and an end-boundary token $\mathbf{e}_e$. Consequently,
$\mathbf{e}_s$ acts as a fixed learned reference for the canonical start
direction, while $\mathbf{e}_e$ carries the varying target-fraction
information. In both tasks, the angles are converted to periodic features
using \cref{eq:periodic_encoding}. The two tasks use the same architecture but
are trained independently with separate projection networks and LoRA
parameters.

The geometric tokens replace placeholder embeddings at
semantically meaningful positions in the prompt before the first transformer
layer of the frozen text encoder. The text encoder and base diffusion-model
weights remain frozen; we train only the projection network $P$ and LoRA
adapters applied to the diffusion denoiser.
Role-specific placement associates each token with a semantic component, but
does not ensure that the token influences the corresponding image region. We
address this spatial-grounding problem next.

\subsection{Localization Relaxation for Grounding}
\label{sec:localization_relaxation}

Role-specific token placement does not by itself guarantee that each token
will affect the intended image region. We therefore build on Coupled Attention
Localization (CALL) from
Compass Control~\cite{rishubh2025compasscontrol}, which constrains the
cross-attention of a geometric token and its associated semantic token to a
specified bounding box.
Providing boxes at inference is natural in Compass Control, where they specify
the locations of independently controlled objects in a multi-object scene. In
our setting, however, boxes for the clock hands or sector boundaries would
partially reveal the geometric configuration that the model is expected to
generate. The model should instead learn where each component belongs from
the target quantity and its semantic role. We therefore use localization only as training-time supervision and
introduce \emph{Stochastic Localization Relaxation (SLR)}. Rather than removing spatial
supervision in a single step, we randomly vary the granularity of spatial supervision through three regimes:

\begin{itemize}[noitemsep,topsep=2pt]

    \item \textbf{Component-level constraint.}
    The cross-attention of each geometric token and its associated semantic
    token is restricted to the bounding box of the corresponding visual
    component. For clocks, the hour- and minute-hand tokens are constrained
    using their respective hand boxes. This establishes a direct spatial
    association between each token and the component it controls.

    \item \textbf{Object-level constraint.}
    The component boxes are replaced by a single bounding box covering the
    complete object. For clocks, the attention of both hand tokens is
    restricted to the full clock region. This identifies the relevant object
    without revealing the positions or orientations of its individual
    components.

    \item \textbf{Unconstrained attention.}
    The localization constraint is removed, and the model uses standard
    cross-attention. The model must determine the relevant regions from the prompt, target quantity, and token semantics alone.

\end{itemize}

The localization type is drawn independently for each training sample from a fixed
categorical distribution with probabilities $55\%$, $30\%$, and $15\%$ for the
component-level, object-level, and unconstrained regimes. These
three regimes span the range from precise component supervision to none at
all, enabling bounding-box-free inference.
At inference, no localization constraint or bounding box is provided. The
model must generate the spatial configuration solely from the text prompt, the target quantity, and the learned semantic roles of the geometric tokens.

\subsection{Hand-Swap Preference Optimization}
\label{sec:hand_swap}

During initial geometry training, we observe that the model occasionally
generates the correct pair of hand angles but assigns them to the wrong hands.
We therefore introduce a clock-specific preference objective that explicitly
favors the correct hour--minute assignment over its swapped counterpart.
Related hand-swapping ambiguities have been used to construct hard negatives
for analog-clock reading~\cite{choi2026clockreasoning}; we adapt this idea to
clock generation with a diffusion-specific objective.
We apply the preference objective in a second training phase initialized from
the geometry-trained model. For each clock sample, the correct conditioning
$\mathbf{c}^{+}$ places the hour- and minute-hand embeddings in their
designated positions in the prompt:

\begin{equation}
    \mathbf{c}^{+}
    =
    \left(
        \mathbf{e}_h,
        \mathbf{e}_m
    \right).
\end{equation}

We construct a hard-negative conditioning by exchanging the two embeddings
while keeping the prompt unchanged:

\begin{equation}
    \mathbf{c}^{-}
    =
    \left(
        \mathbf{e}_m,
        \mathbf{e}_h
    \right).
    \label{eq:swapped_conditioning}
\end{equation}

The correct and swapped conditionings contain the same angular information
but associate it with different semantic roles. We evaluate both using the
same diffusion timestep, noisy latent $\mathbf{z}_t$, and noise target
$\boldsymbol{\epsilon}$, ensuring that their difference reflects the
hour--minute assignment.
For a conditioning sequence $\mathbf{c}$, we define the denoising error of the
trainable model as

\begin{equation}
    \ell_{\theta}(\mathbf{c})
    =
    \left\|
        \boldsymbol{\epsilon}
        -
        \boldsymbol{\epsilon}_{\theta}
        \left(
            \mathbf{z}_t,
            t,
            \mathbf{c}
        \right)
    \right\|_2^2.
    \label{eq:conditioning_error}
\end{equation}

Let $\ell_{\mathrm{ref}}(\mathbf{c})$ denote the corresponding error under a
fixed reference branch. In this branch, the projection network is fixed at
its geometry-trained parameters and the U-Net LoRA contribution is disabled.
We measure improvement relative to this reference as

\begin{equation}
    r(\mathbf{c})
    =
    \ell_{\mathrm{ref}}(\mathbf{c})
    -
    \ell_{\theta}(\mathbf{c}).
    \label{eq:preference_reward}
\end{equation}

We then optimize the pairwise preference loss

\begin{equation}
    \mathcal{L}_{\mathrm{swap}}
    =
    -\log
    \sigma\!\left(
        \beta
        \left[
            r(\mathbf{c}^{+})
            -
            r(\mathbf{c}^{-})
        \right]
    \right),
    \label{eq:swap_loss}
\end{equation}

where $\sigma$ is the logistic sigmoid and $\beta$ controls the preference
strength. The objective encourages greater improvement over the reference
under the correct conditioning than under the swapped conditioning.
The final clock-training objective is

\begin{equation}
    \mathcal{L}
    =
    \mathcal{L}_{\mathrm{diff}}
    +
    \lambda_{\mathrm{swap}}
    \mathcal{L}_{\mathrm{swap}},
    \label{eq:clock_objective}
\end{equation}

where $\mathcal{L}_{\mathrm{diff}}$ is the standard diffusion denoising loss.
The hand-swap preference objective is applied only to the clock model; the
fraction model is trained using $\mathcal{L}_{\mathrm{diff}}$ alone.

\section{Experiments}
\label{sec:experiments}

\noindent\textbf{Training data.}
\begin{figure}[t]
  \centering
  \includegraphics[width=\linewidth]{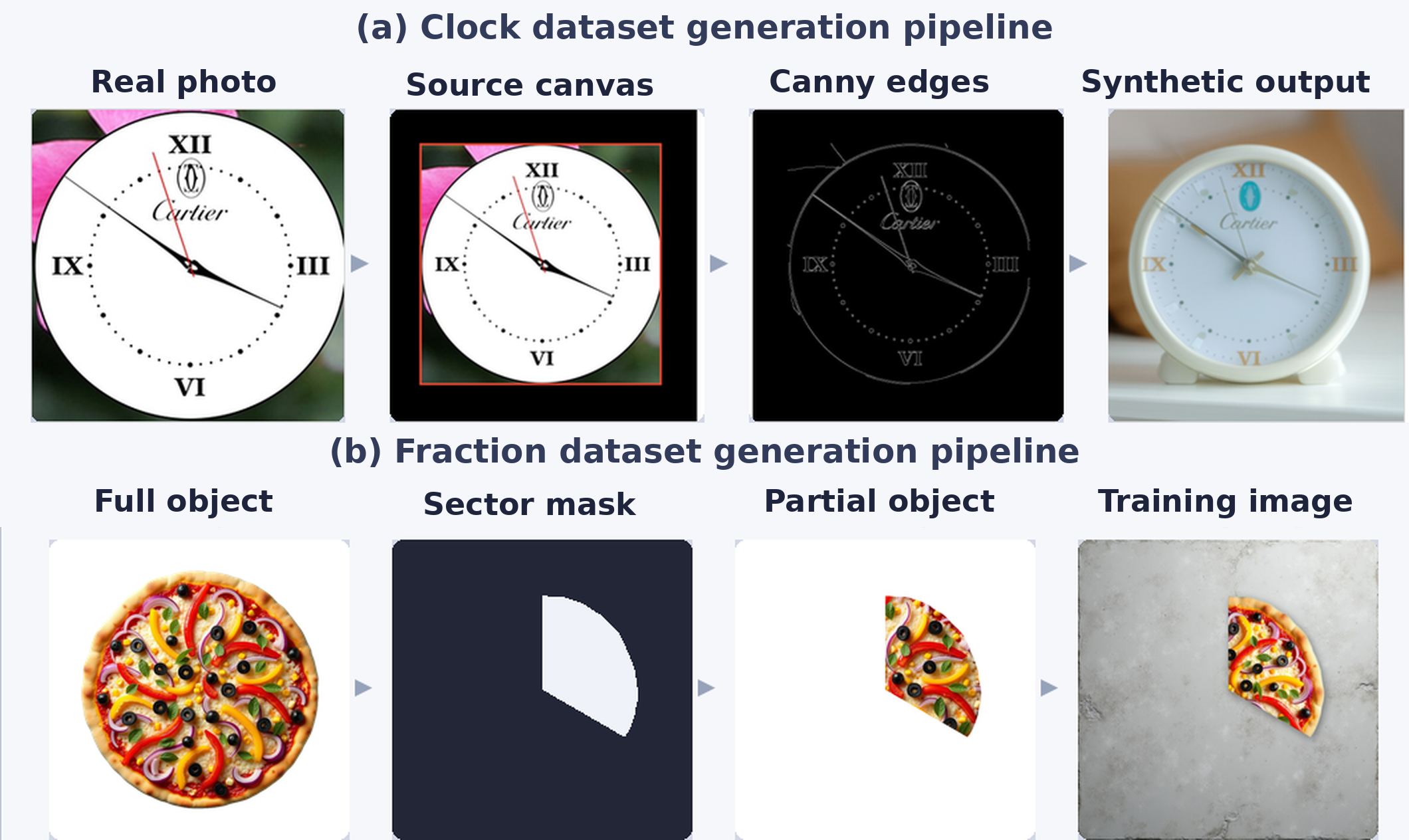}
  \caption{
    Dataset construction for the two domains.
    \textbf{Top:} Procedurally rendered clocks are restyled using
    ControlNet conditioned on Canny edges while preserving exact hand angles.
    \textbf{Bottom:} Complete circular objects generated by Flux.1 are
    segmented and arc-masked according to the target fraction.
  }
  \label{fig:datasets}
\end{figure}
We construct separate training sets for clock-time and circular-fraction
generation, as illustrated in Figure~\ref{fig:datasets}. For clocks, we begin
with procedurally rendered analog clocks covering the full 12-hour cycle with
precise hand-angle annotations. We extract Canny edges and use a Stable
Diffusion~3.5 ControlNet~\cite{esser2024scaling,zhang2023adding} to generate
photorealistic variants across 12 visual styles while preserving the underlying
hand geometry.

For fractions, we synthesize 10{,}000 images over 32 circular object
categories and 72 angular extents from $5^\circ$ to $360^\circ$ in
$5^\circ$ increments. Flux.1~\cite{flux2024} generates complete circular
objects; we then segment each object, apply the target arc mask, and composite
it onto diverse backgrounds. Appendix~A gives full implementation details.

\begin{table*}[t]
\centering
\footnotesize
\caption{Clock generation results under three automatic readers,
human annotation, and a user study with 33 participants.
Acc@$6^\circ$: both hand errors $\leq 6^\circ$.
Avg. MAE: mean angular error averaged over both hands.
The user study reports the percentage of selections for
geometric controllability (Ctrl.) and appearance quality (App.).
Additional times: 48 non-quarter-hour targets (960 images).}
\label{tab:clock_results}

\setlength{\tabcolsep}{2.5pt}
\begin{tabular}{lcccccccccc}
\toprule
& \multicolumn{2}{c}{\textbf{GPT-5.6}}
& \multicolumn{2}{c}{\textbf{Claude Opus 5}}
& \multicolumn{2}{c}{\textbf{IAT}}
& \multicolumn{2}{c}{\textbf{Human}}
& \multicolumn{2}{c}{\textbf{User Study}} \\
\cmidrule(lr){2-3}
\cmidrule(lr){4-5}
\cmidrule(lr){6-7}
\cmidrule(lr){8-9}
\cmidrule(lr){10-11}

\textbf{Model}
& A@$6^\circ$$\uparrow$ & Avg. MAE$\downarrow$
& A@$6^\circ$$\uparrow$ & Avg. MAE$\downarrow$
& A@$6^\circ$$\uparrow$ & Avg. MAE$\downarrow$
& A@$6^\circ$$\uparrow$ & Avg. MAE$\downarrow$
& Ctrl.$\uparrow$ & App.$\uparrow$ \\
\midrule

SD 2.1~\cite{rombach2022highresolution}
& 0.4\% & $88.1^\circ$
& 0.9\% & $91.3^\circ$
& 0.5\% & $89.7^\circ$
& 0.0\% & $104.5^\circ$
& 1.5\% & 5.7\% \\

SDXL~\cite{podell2024sdxl}
& 0.2\% & $90.7^\circ$
& 0.7\% & $91.2^\circ$
& 0.4\% & $90.7^\circ$
& 0.0\% & $112.8^\circ$
& 1.3\% & 7.4\% \\

SD 3.5~\cite{esser2024scaling}
& 0.0\% & $88.4^\circ$
& 0.0\% & $87.8^\circ$
& 0.1\% & $90.7^\circ$
& 0.0\% & $89.8^\circ$
& 1.9\% & 9.7\% \\

Flux.1~\cite{flux2024}
& 0.0\% & $91.9^\circ$
& 0.0\% & $87.9^\circ$
& 0.6\% & $90.4^\circ$
& 0.0\% & $92.6^\circ$
& 5.7\% & 26.9\% \\

\midrule

Ours (train)
& \textbf{71.9\%} & $\mathbf{21.0^\circ}$
& \textbf{63.4\%} & $\mathbf{23.6^\circ}$
& \textbf{76.6\%} & $\mathbf{19.9^\circ}$
& \multirow{2}{*}{\textbf{75.3\%}}
& \multirow{2}{*}{$\mathbf{20.7^\circ}$}
& \multirow{2}{*}{\textbf{89.6\%}}
& \multirow{2}{*}{\textbf{50.3\%}} \\

Ours (unseen)
& \textbf{71.6\%} & $\mathbf{22.9^\circ}$
& \textbf{62.8\%} & $\mathbf{25.1^\circ}$
& \textbf{79.4\%} & $\mathbf{17.2^\circ}$
& & & & \\

\midrule

Ours (additional times)
& \textbf{50.7\%} & $\mathbf{28.5^\circ}$
& \textbf{46.7\%} & $\mathbf{33.8^\circ}$
& \textbf{77.6\%} & $\mathbf{18.1^\circ}$
&  & 
&  &  \\

\bottomrule
\end{tabular}
\end{table*}

\begin{table}[t]
\centering
\footnotesize
\caption{Fraction generation results. Acc@$k$: fraction error $\leq k$.
Baselines are evaluated without fine-tuning. Unseen objects:
14 novel categories. Unseen fractions: held-out fraction values.
Dense $360^\circ$: all $1^\circ$ steps over 3 objects.
User study reports controllability/appearance preference (\%).}
\label{tab:fraction_results}

\setlength{\tabcolsep}{2.5pt}
\begin{tabular}{l c c c c c}
\toprule
\textbf{Method / Split}
& \textbf{MAE}$\downarrow$
& \textbf{A@.05}$\uparrow$
& \textbf{A@.10}$\uparrow$
& \textbf{Spearman}$\uparrow$
& \textbf{User}$\uparrow$ \\
\midrule

SD~2.1~\cite{rombach2022highresolution}
& 0.288 & 10.4\% & 20.1\% & $-$0.005
& 2.5/4.5 \\

SDXL~\cite{podell2024sdxl}
& 0.293 & 10.8\% & 20.6\% & 0.012
& 1.1/5.1 \\

SD~3.5~\cite{esser2024scaling}
& 0.305 & 10.4\% & 19.5\% & 0.003
& 1.8/6.8 \\

Flux.1~\cite{flux2024}
& 0.308 & 9.6\% & 19.9\% & 0.004
& 1.7/16.2 \\

\midrule

Ours (train obj.)
& 0.065
& 82.0\%
& \textbf{89.5\%}
& \textbf{0.821}
& \multirow{4}{*}{\textbf{92.8/67.4}} \\

Ours (unseen obj.)
& \textbf{0.064} & \textbf{84.1\%} & 88.8\% & 0.797 & \\

Ours (unseen frac.)
& 0.075 & 80.5\% & 84.9\% & 0.805 & \\

Ours (dense $360^\circ$)
& 0.091 & 74.3\% & 81.6\% & 0.726 & \\

\bottomrule
\end{tabular}
\end{table}

\noindent\textbf{Evaluation protocol.}
For clock generation, we evaluate all 48 quarter-hour targets using 20 style
prompts per target, including both training and unseen styles. We use three independent automatic clock readers: GPT-5.6, Claude Opus~5,
and ItsAboutTime (IAT)~\cite{yang2021its}, a ResNet50-based
clock reader trained on synthetic and pseudo-labeled real data.
The two vision-language models (VLMs) were selected based on their performance on the ClockBench analog-clock reading benchmark~\cite{clockbench}. Because VLMs are known to be unreliable on precise geometric perception~\cite{rahmanzadehgervi2024blind},
we corroborate them with IAT and human annotation rather than relying on
any single reader. We also conduct a human evaluation in
which three annotators independently read the displayed time from a subset of
generated clocks. Predicted \texttt{HH:MM} values from the automatic readers
and human annotators are converted to hour- and minute-hand angles and compared
with the requested time using circular angular distance. We report
\textbf{Acc@$6^\circ$}, where both hand errors must be within $6^\circ$,
together with the average
mean absolute circular error (\textbf{Avg. MAE})
across the hour and minute hands.
We further conduct a user study with 33 participants, who select the best output
for controllability and appearance among anonymized model generations.

We evaluate 71 target fractions with angular extents from
$5^\circ$ to $355^\circ$ in $5^\circ$ increments. A vision-based detector
estimates the visible sector angle clockwise from 12~o'clock and converts
it to a fraction in $[0,1]$. We report \textbf{Acc@0.10}, the fraction of
predictions within $\pm0.10$ of the target, and \textbf{Spearman $\rho$}
between requested and predicted fractions, separately for train-like and
unseen object categories. Additional implementation and evaluation details are provided in Appendix~D.

\begin{figure}[!t]
\centering

\newlength{\Ccw}
\setlength{\Ccw}{1.35cm}
\newlength{\Crw}
\setlength{\Crw}{0.50cm}

\newcommand{\Cig}[1]{%
  \includegraphics[
    width=\Ccw,
    height=\Ccw,
    keepaspectratio=false
  ]{figs/cmp/#1}%
}
\newcommand{\Crlab}[1]{%
  \parbox[b][\Ccw][c]{\Crw}{%
    \centering\scriptsize #1%
  }%
}
\newcommand{\Chead}[1]{%
  \makebox[\Ccw][c]{\scriptsize\bfseries #1}%
}

\setlength{\tabcolsep}{0.5pt}
\renewcommand{\arraystretch}{1}

\resizebox{\linewidth}{!}{%
\begin{tabular}{@{}r@{\hspace{4pt}}cccc@{\hspace{3pt}\vrule width 0.4pt\hspace{3pt}}c@{}}
  & \multicolumn{4}{c}{\small Baselines} & \small Ours \\[-1pt]
  & \Chead{SD~2.1}
  & \Chead{SDXL}
  & \Chead{SD~3.5}
  & \Chead{Flux.1}
  & \Chead{Ours} \\[2pt]

  \Crlab{01:45}
  & \Cig{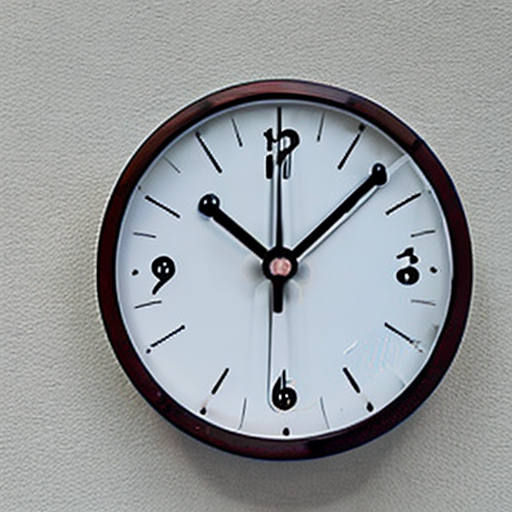}
  & \Cig{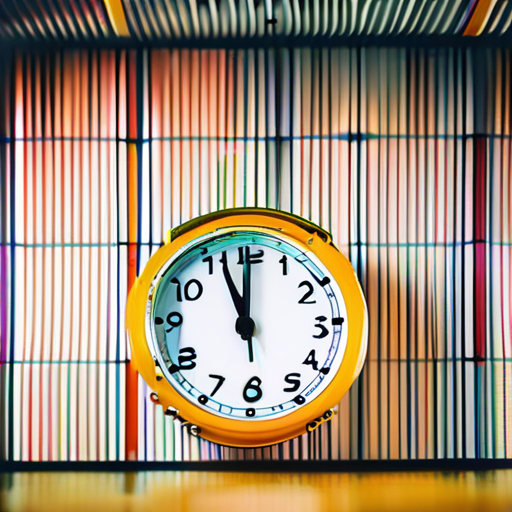}
  & \Cig{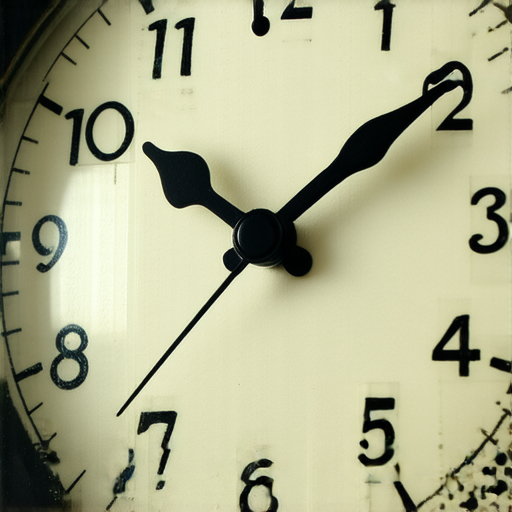}
  & \Cig{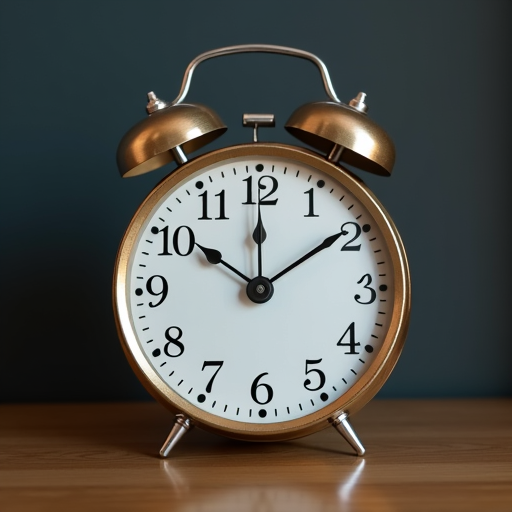}
  & \Cig{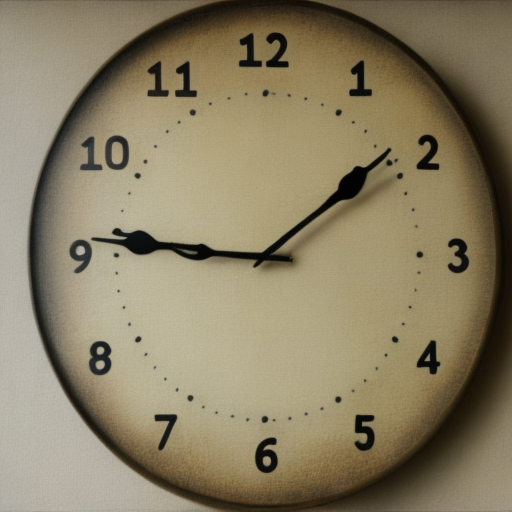} \\

  \Crlab{04:30}
  & \Cig{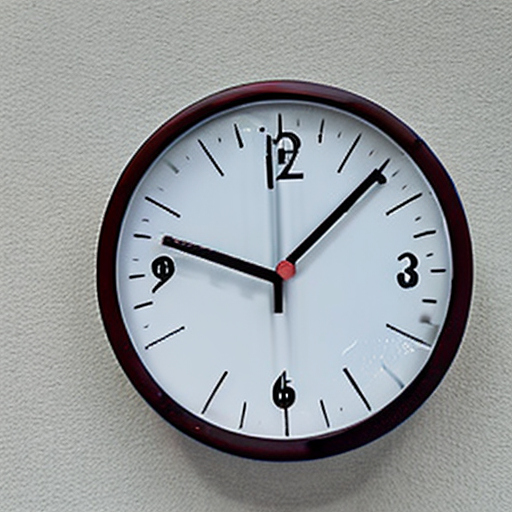}
  & \Cig{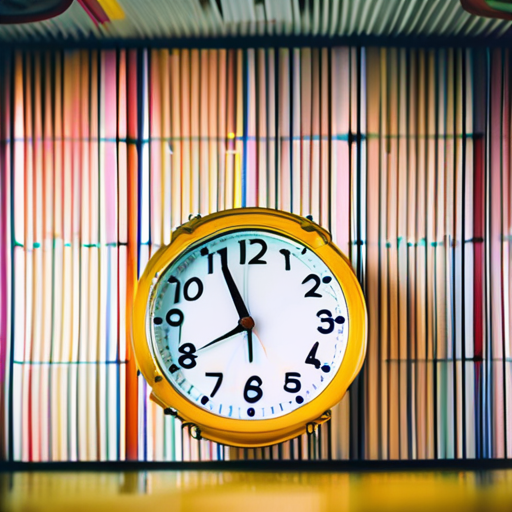}
  & \Cig{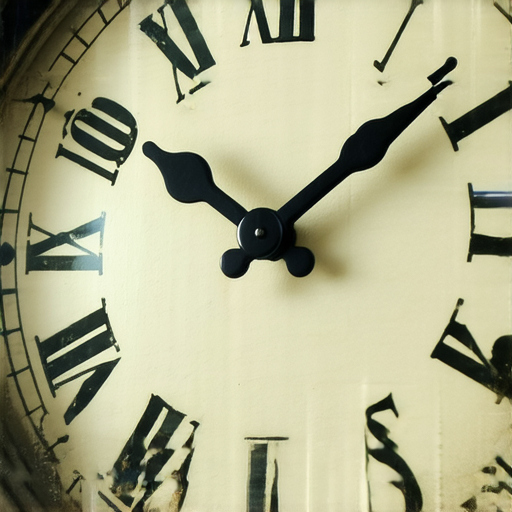}
  & \Cig{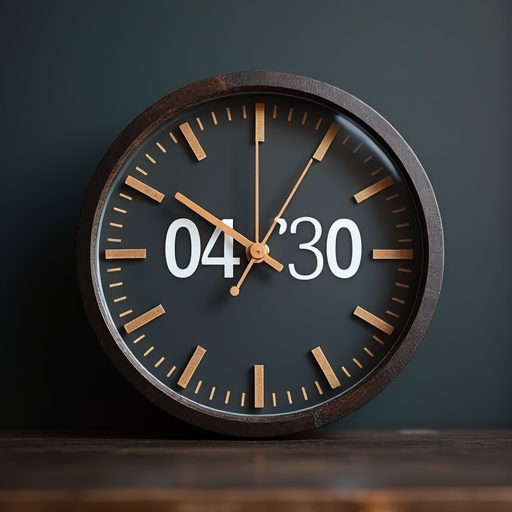}
  & \Cig{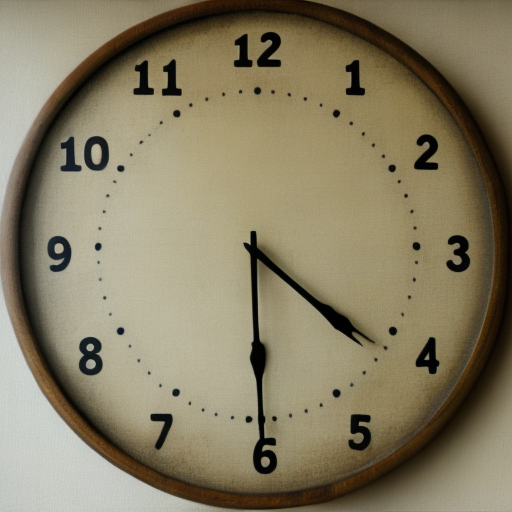} \\

  \Crlab{07:15}
  & \Cig{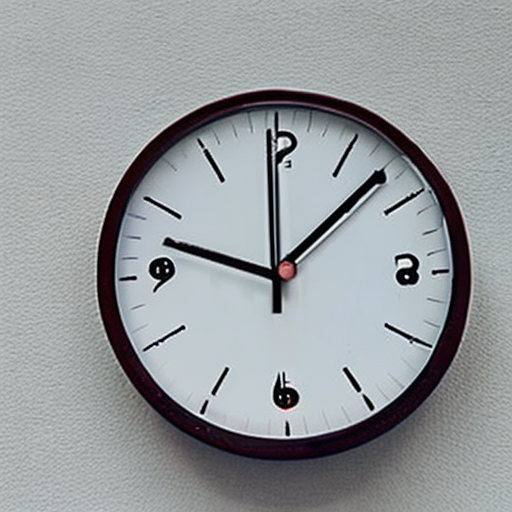}
  & \Cig{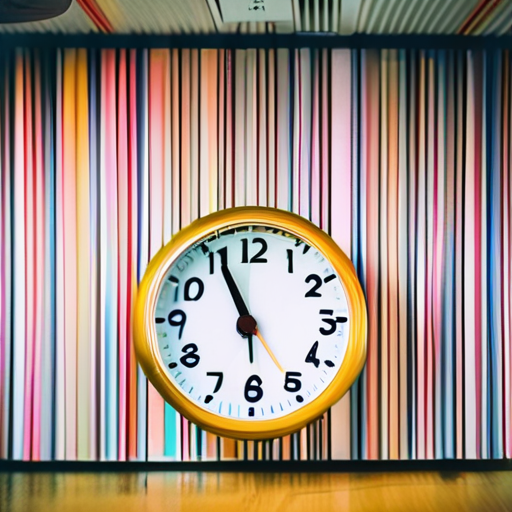}
  & \Cig{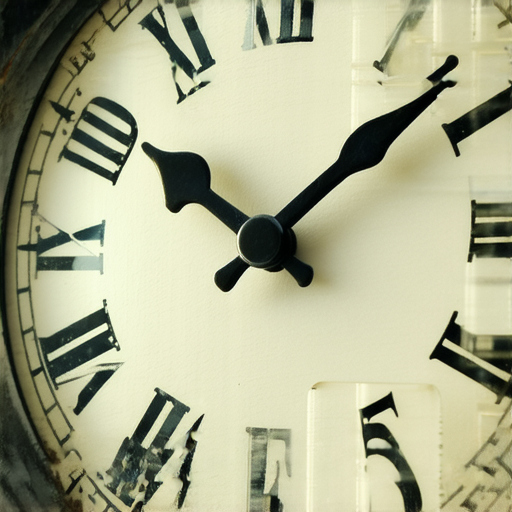}
  & \Cig{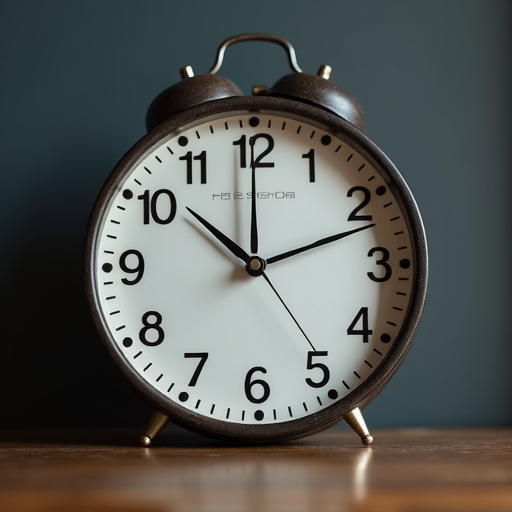}
  & \Cig{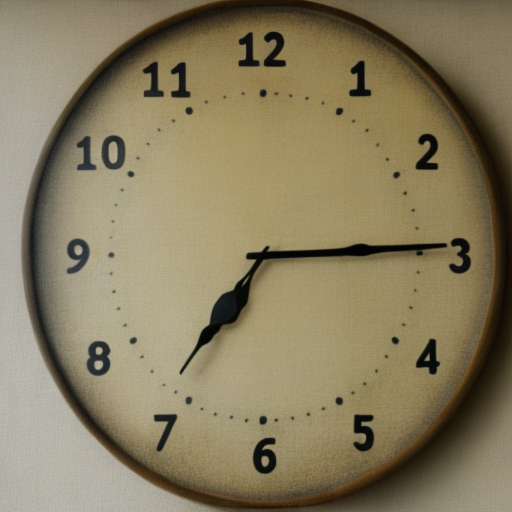} \\

  \Crlab{10:00}
  & \Cig{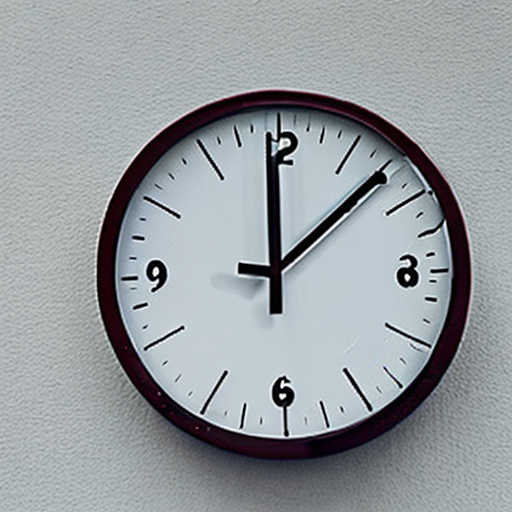}
  & \Cig{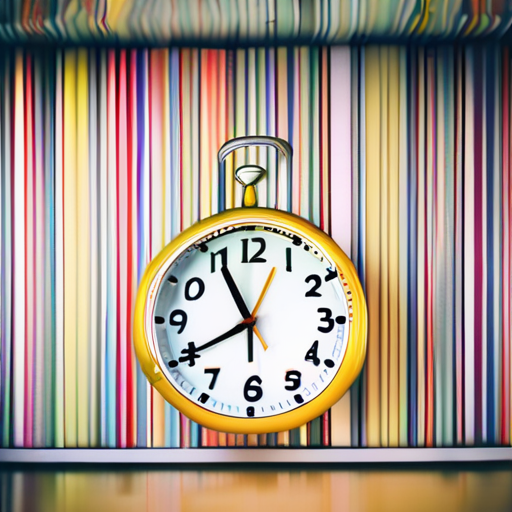}
  & \Cig{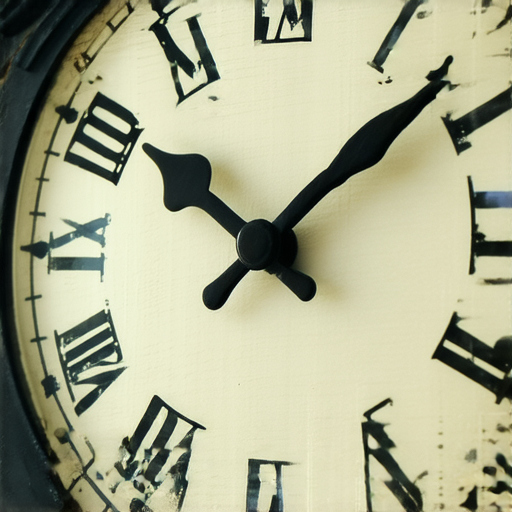}
  & \Cig{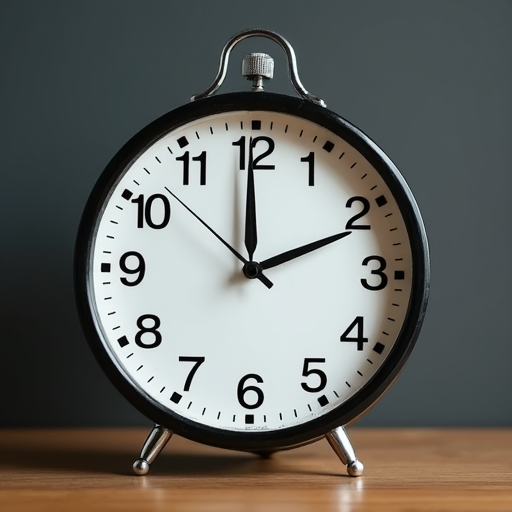}
  & \Cig{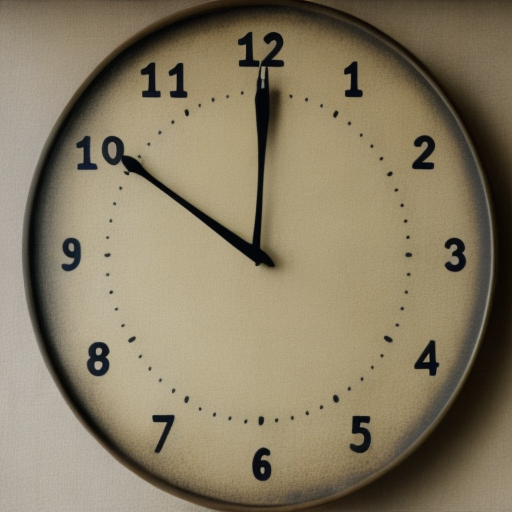} \\
\end{tabular}%
}

\caption{
\textbf{Clock generation.}
Under the same random seeds, baseline models often produce common clock configurations instead of the requested time, whereas our model accurately follows the target time.
}
\label{fig:clock_comparison}
\end{figure}

\begin{figure*}[!t]
\centering

\newlength{\Fcw}
\setlength{\Fcw}{1.40cm}
\newlength{\Frw}
\setlength{\Frw}{0.45cm}

\newcommand{\Figi}[1]{%
  \includegraphics[
    width=\Fcw,
    height=\Fcw,
    keepaspectratio=false
  ]{figs/cmp/#1}%
}
\newcommand{\Frlab}[1]{%
  \parbox[b][\Fcw][c]{\Frw}{%
    \centering\footnotesize #1%
  }%
}
\newcommand{\Fhead}[1]{%
  \makebox[\Fcw][c]{\footnotesize\bfseries #1}%
}

\setlength{\tabcolsep}{1pt}
\renewcommand{\arraystretch}{1}

\begin{minipage}[t]{0.49\linewidth}
\centering
\begin{tabular}{@{}r@{\hspace{2pt}}cccc@{\hspace{4pt}\vrule width 0.4pt\hspace{4pt}}c@{}}
  & \multicolumn{5}{c}{\small\bfseries Pizza} \\[1pt]
  & \Fhead{SD~2.1}
  & \Fhead{SDXL}
  & \Fhead{SD~3.5}
  & \Fhead{Flux.1}
  & \Fhead{Ours} \\[2pt]

  \Frlab{1/5}
  & \Figi{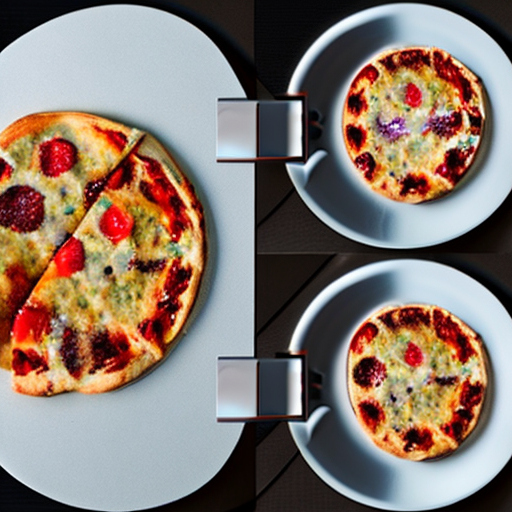}
  & \Figi{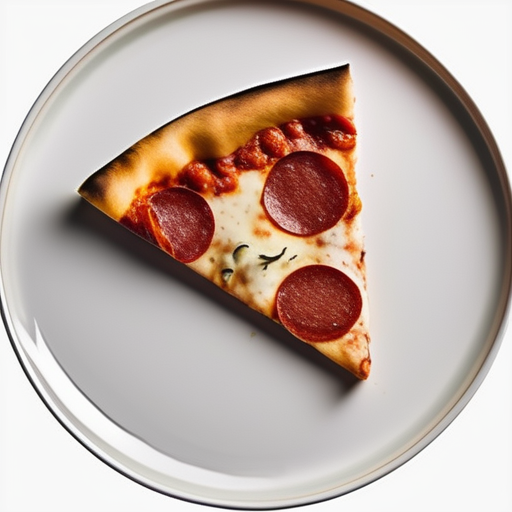}
  & \Figi{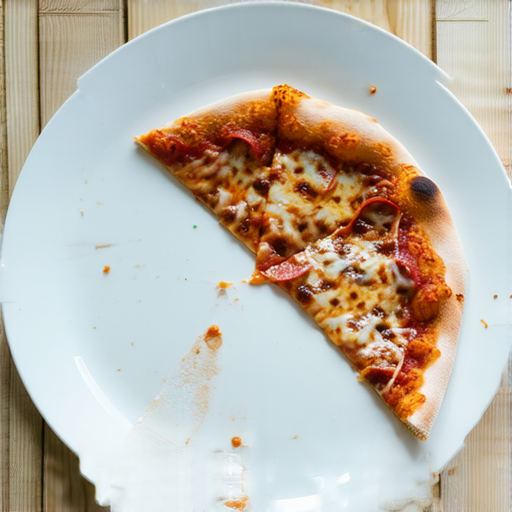}
  & \Figi{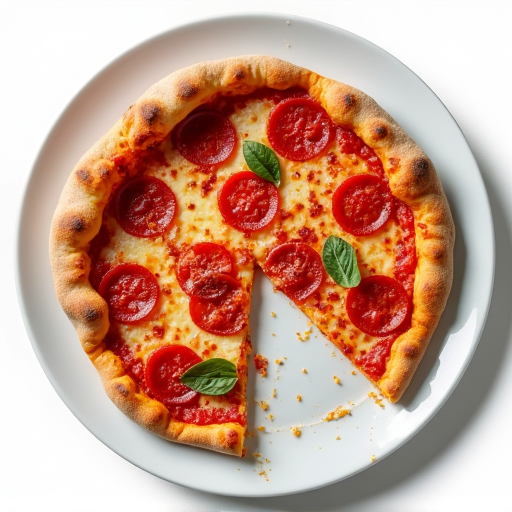}
  & \Figi{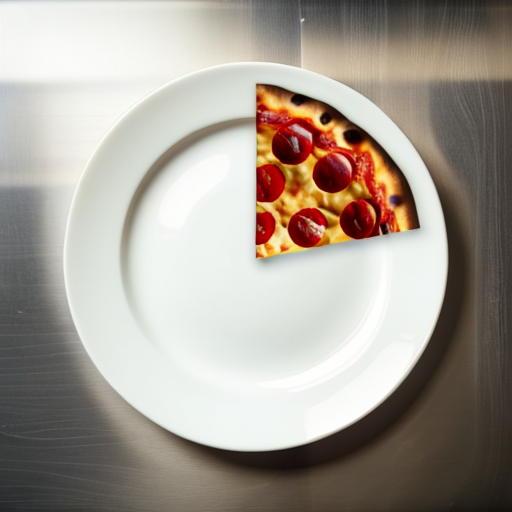} \\

  \Frlab{1/3}
  & \Figi{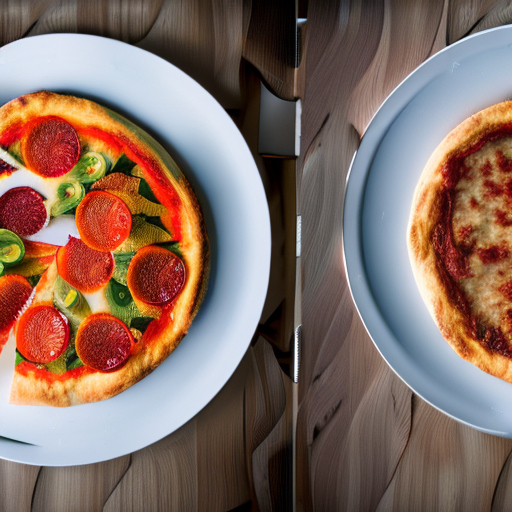}
  & \Figi{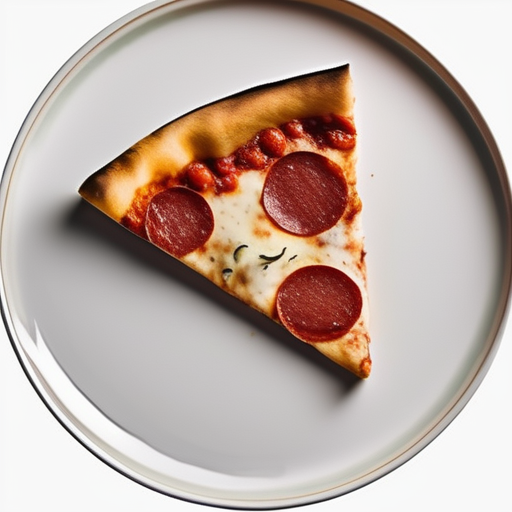}
  & \Figi{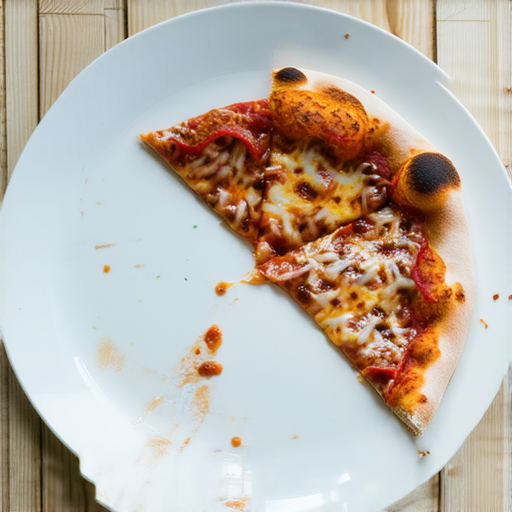}
  & \Figi{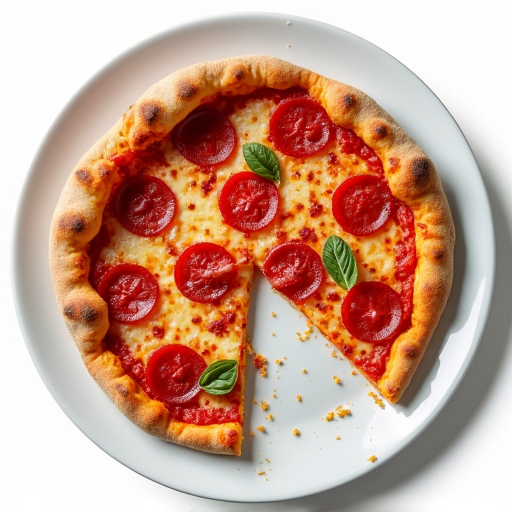}
  & \Figi{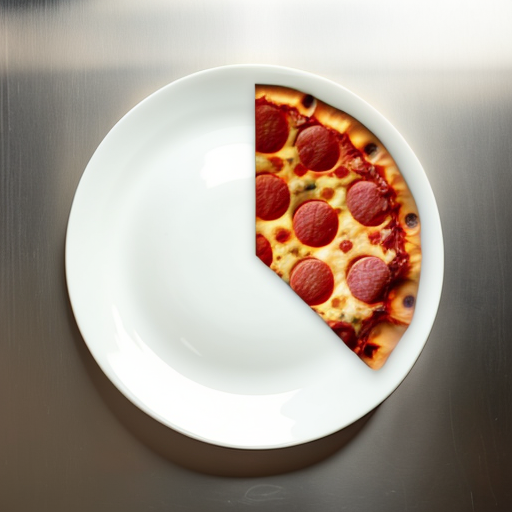} \\

  \Frlab{1/2}
  & \Figi{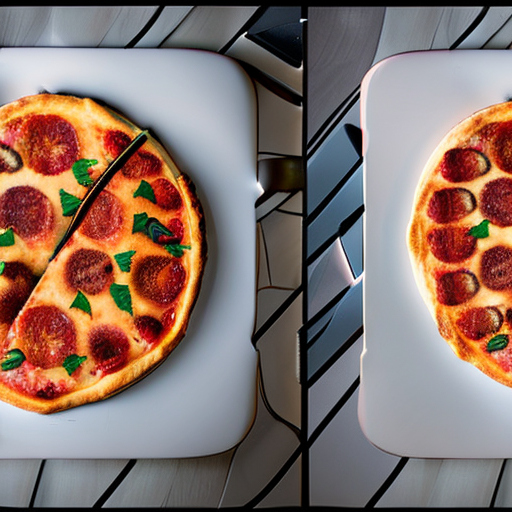}
  & \Figi{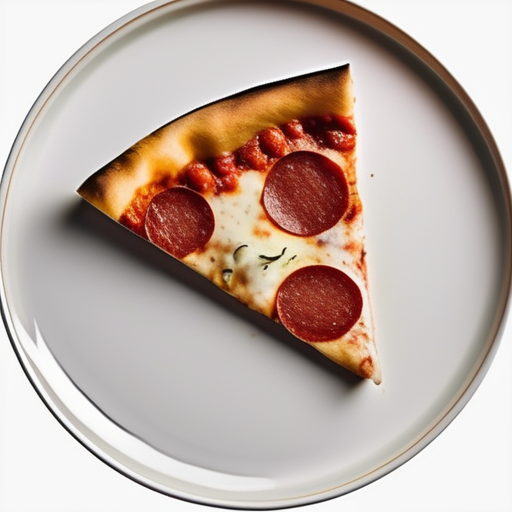}
  & \Figi{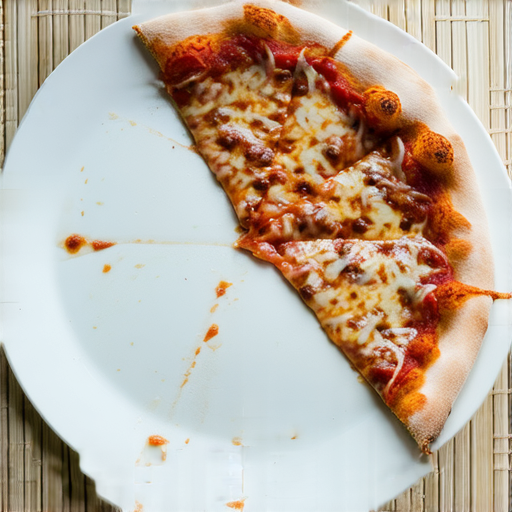}
  & \Figi{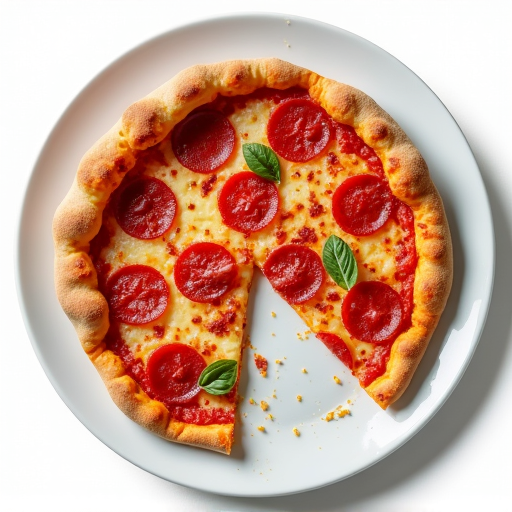}
  & \Figi{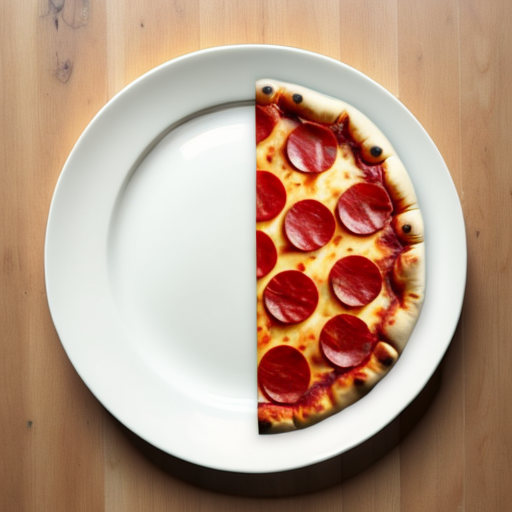} \\

  \Frlab{2/3}
  & \Figi{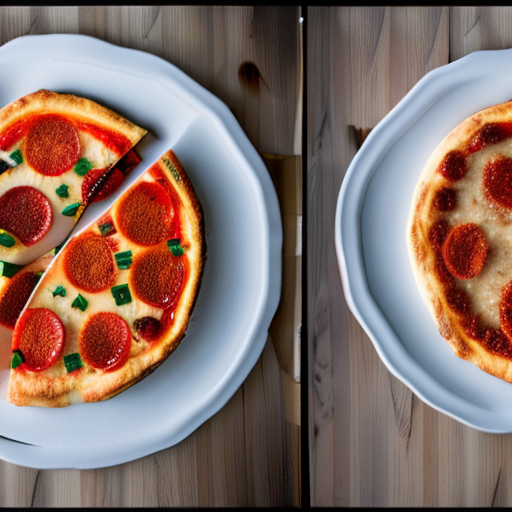}
  & \Figi{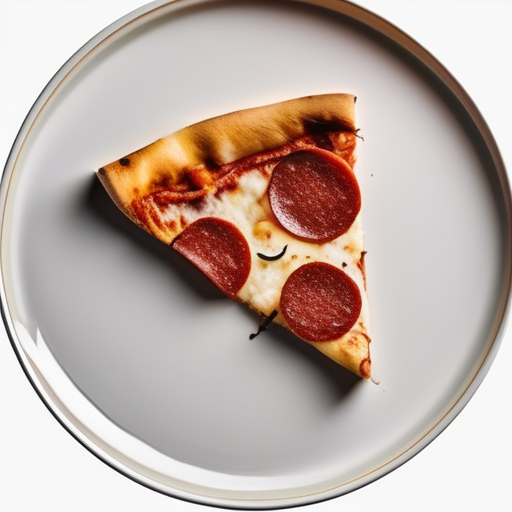}
  & \Figi{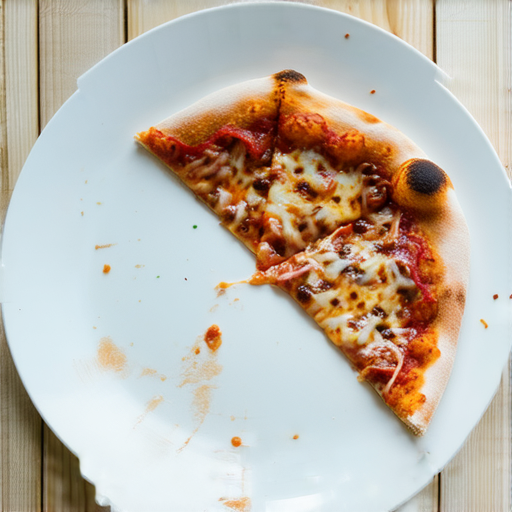}
  & \Figi{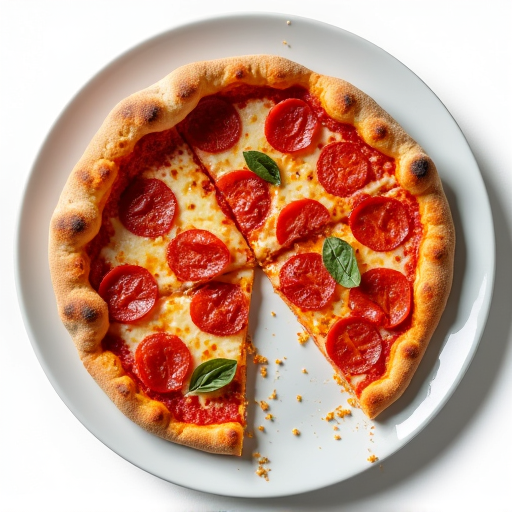}
  & \Figi{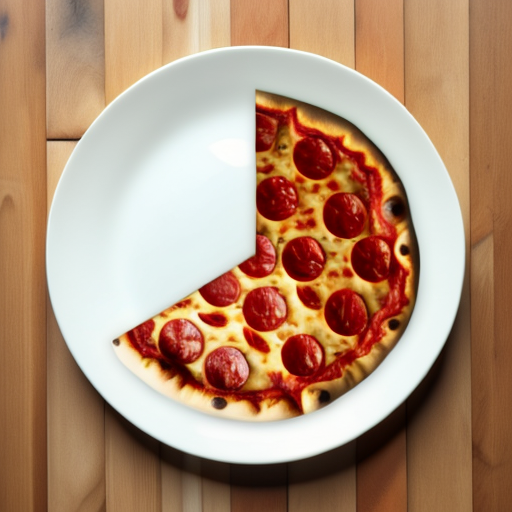} \\

  \Frlab{4/5}
  & \Figi{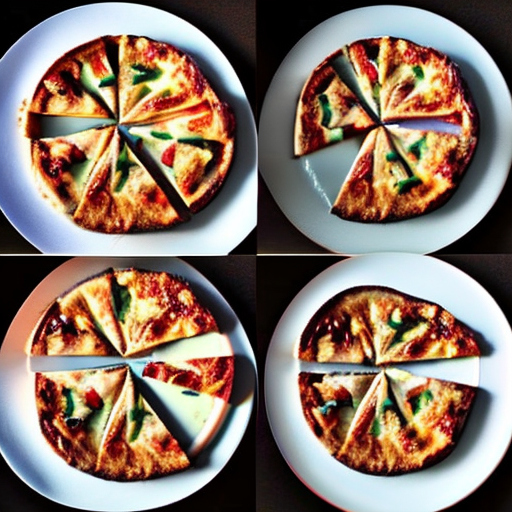}
  & \Figi{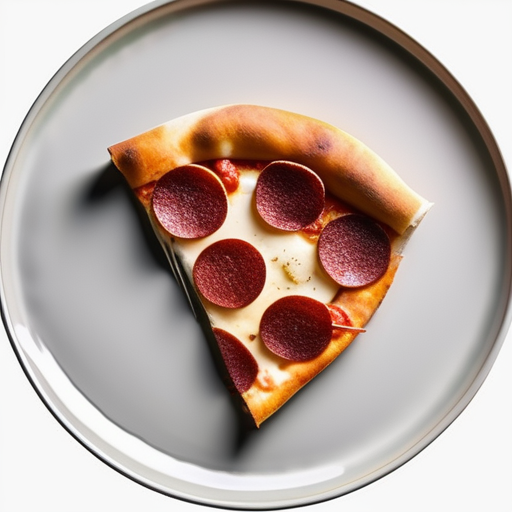}
  & \Figi{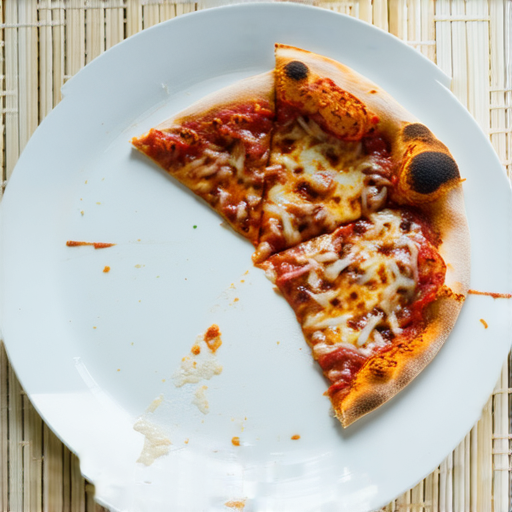}
  & \Figi{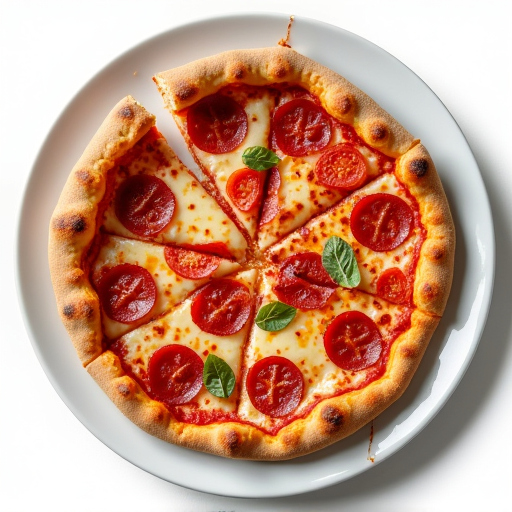}
  & \Figi{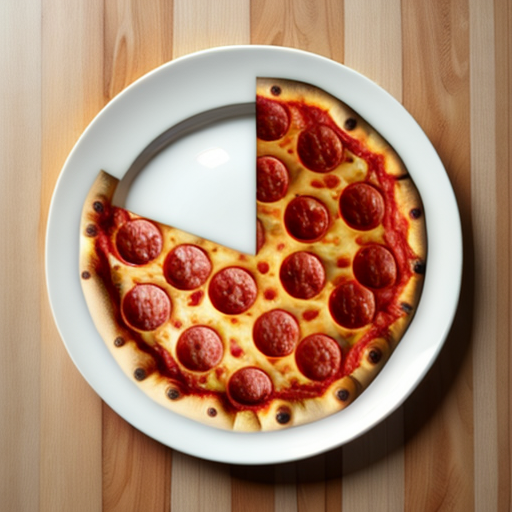} \\
\end{tabular}
\end{minipage}%
\hfill
\begin{minipage}[t]{0.49\linewidth}
\centering
\begin{tabular}{@{}r@{\hspace{2pt}}cccc@{\hspace{4pt}\vrule width 0.4pt\hspace{4pt}}c@{}}
  & \multicolumn{5}{c}{\small\bfseries Pie} \\[1pt]
  & \Fhead{SD~2.1}
  & \Fhead{SDXL}
  & \Fhead{SD~3.5}
  & \Fhead{Flux.1}
  & \Fhead{Ours} \\[2pt]

  \Frlab{1/5}
  & \Figi{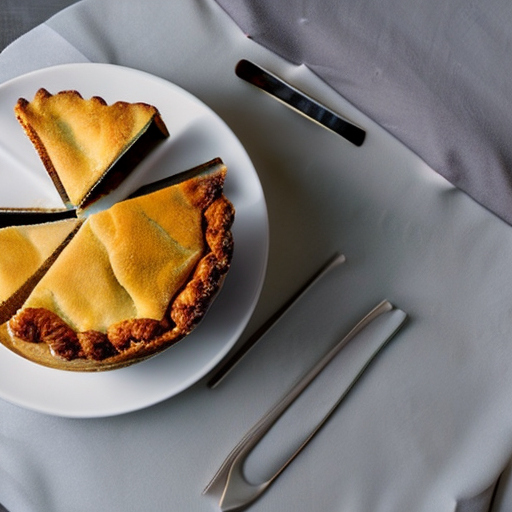}
  & \Figi{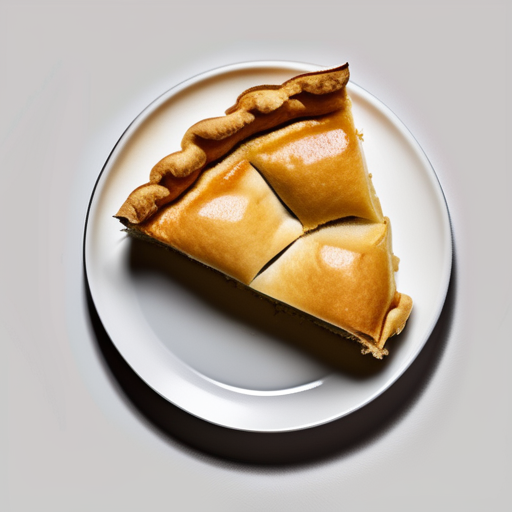}
  & \Figi{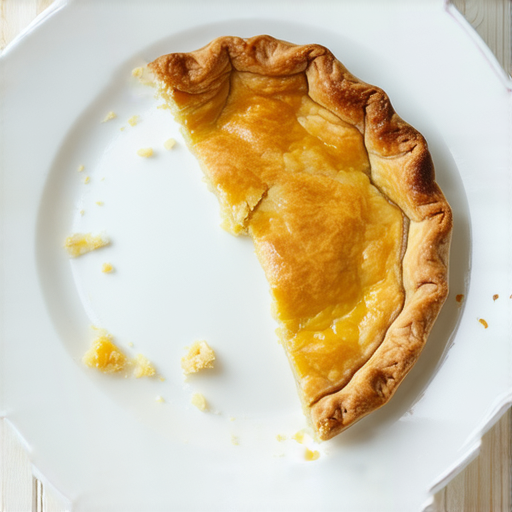}
  & \Figi{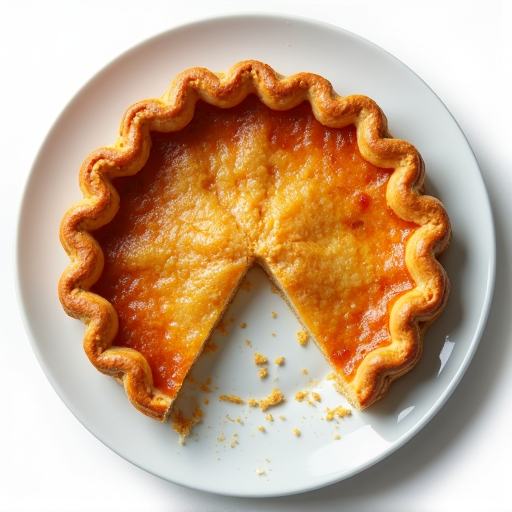}
  & \Figi{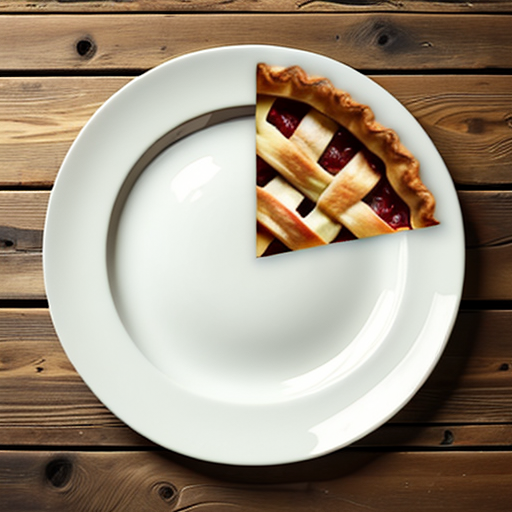} \\

  \Frlab{1/3}
  & \Figi{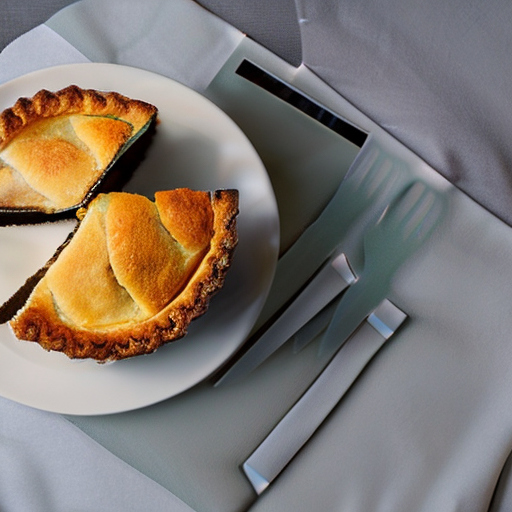}
  & \Figi{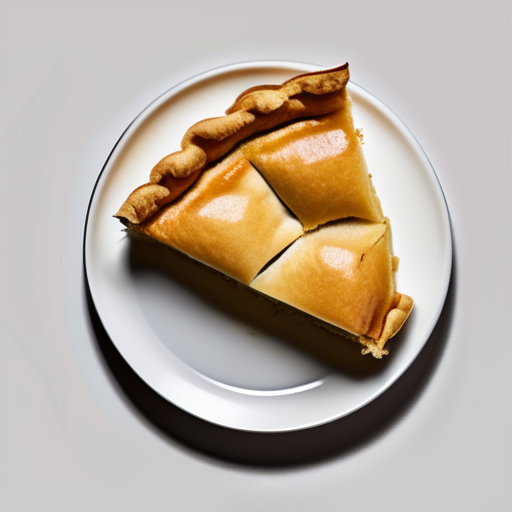}
  & \Figi{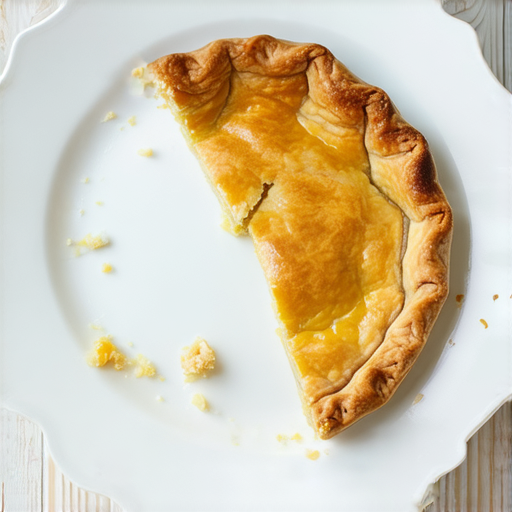}
  & \Figi{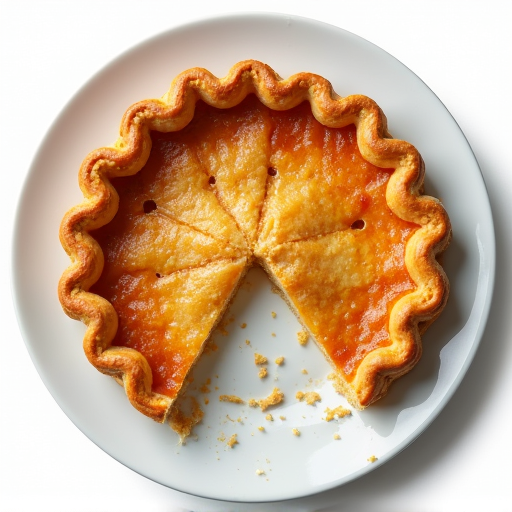}
  & \Figi{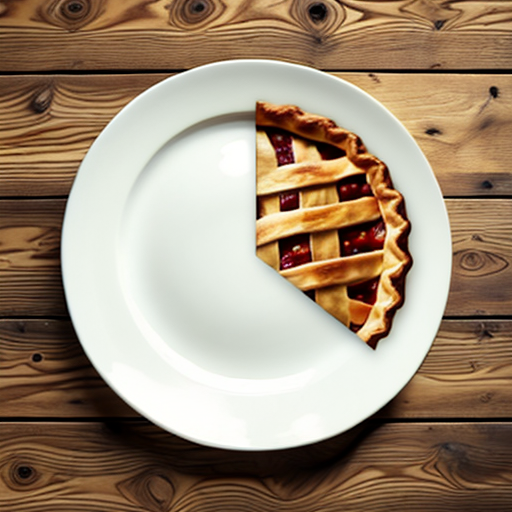} \\

  \Frlab{1/2}
  & \Figi{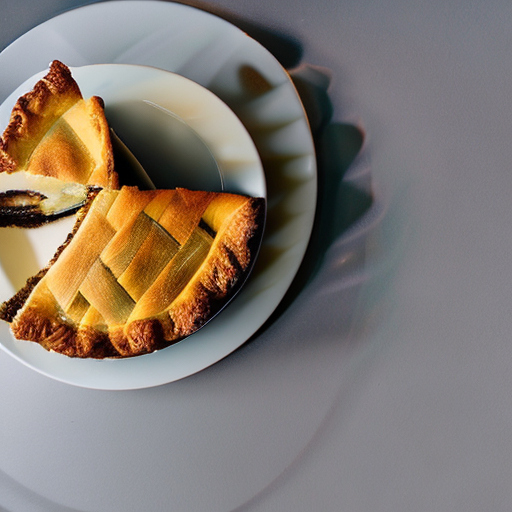}
  & \Figi{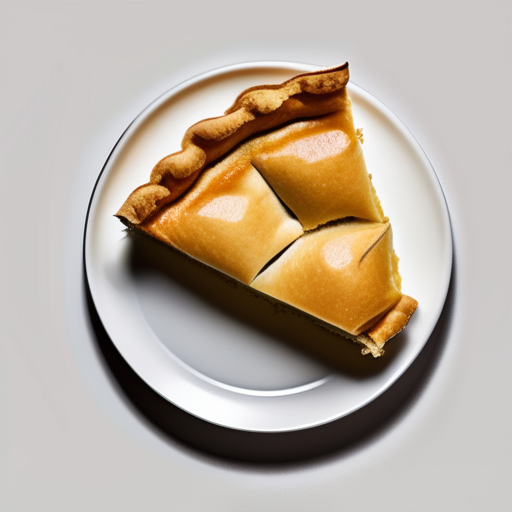}
  & \Figi{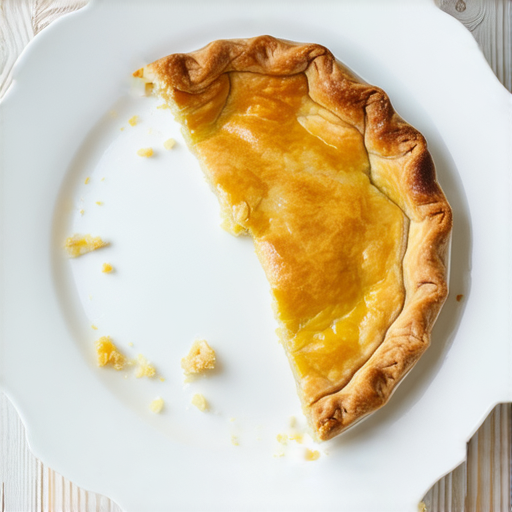}
  & \Figi{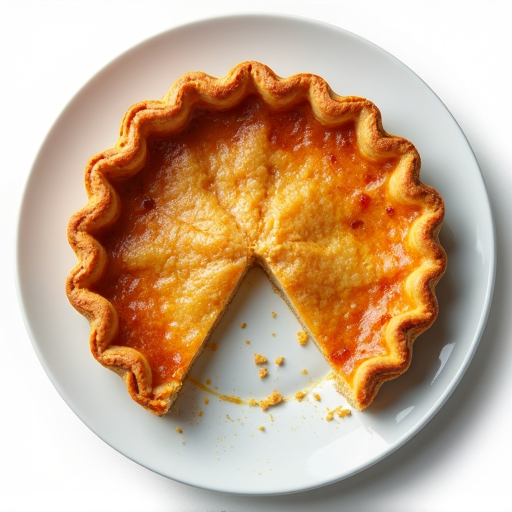}
  & \Figi{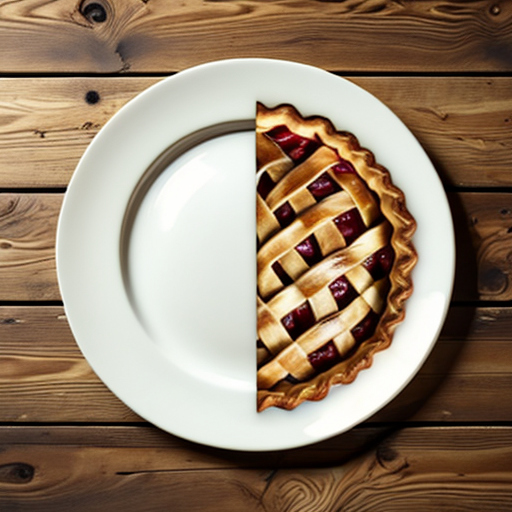} \\

  \Frlab{2/3}
  & \Figi{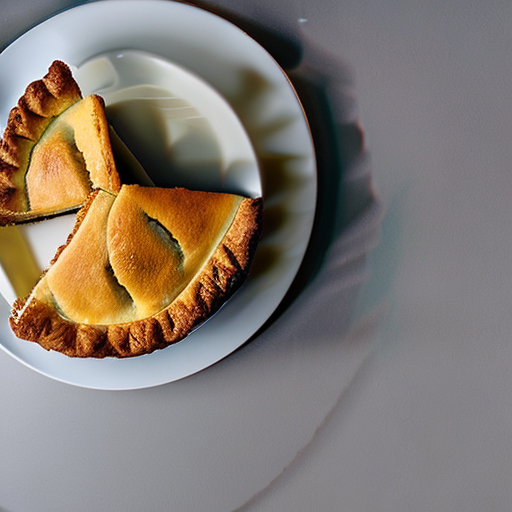}
  & \Figi{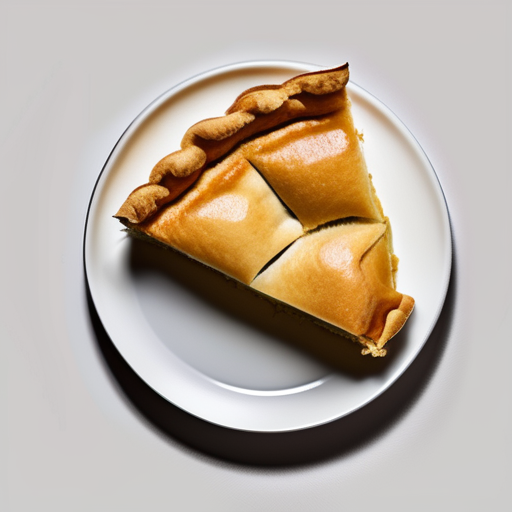}
  & \Figi{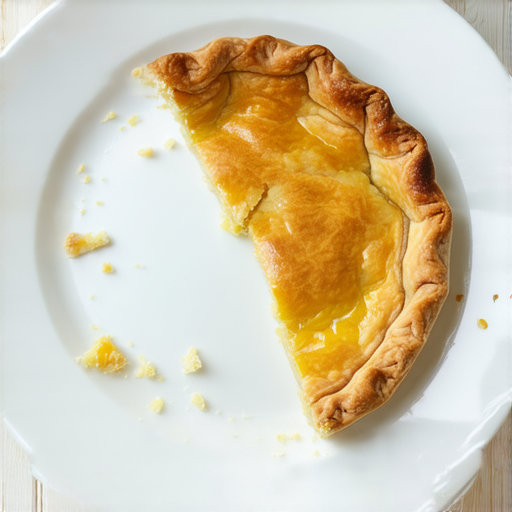}
  & \Figi{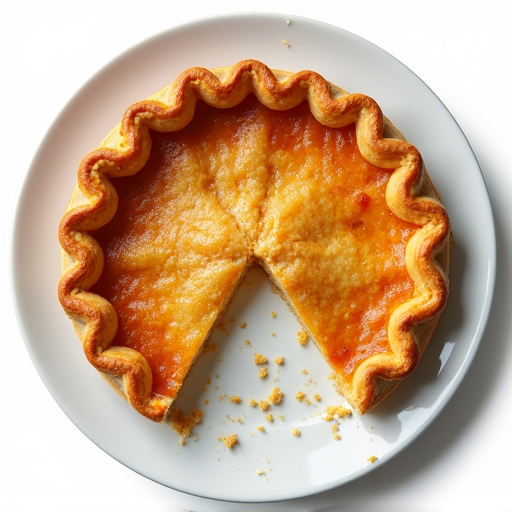}
  & \Figi{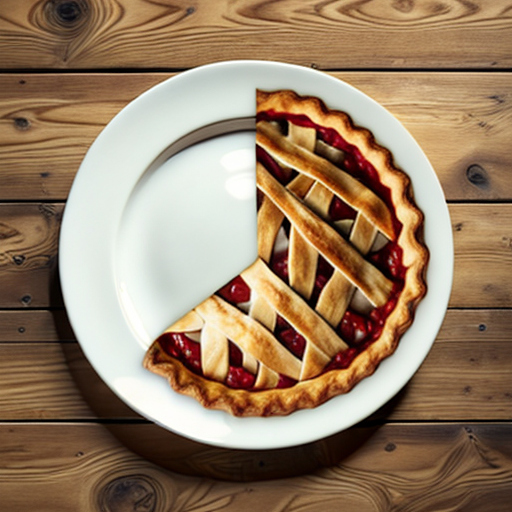} \\

  \Frlab{4/5}
  & \Figi{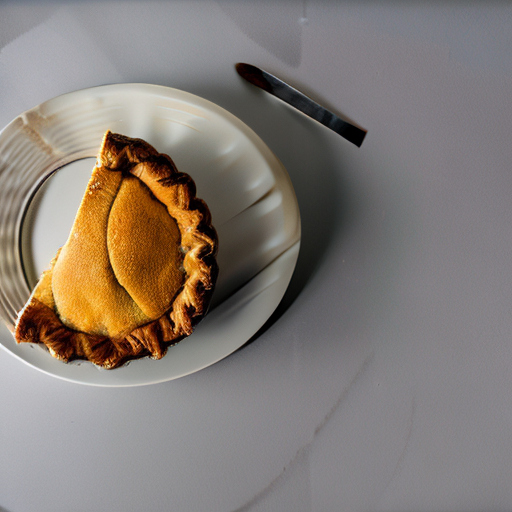}
  & \Figi{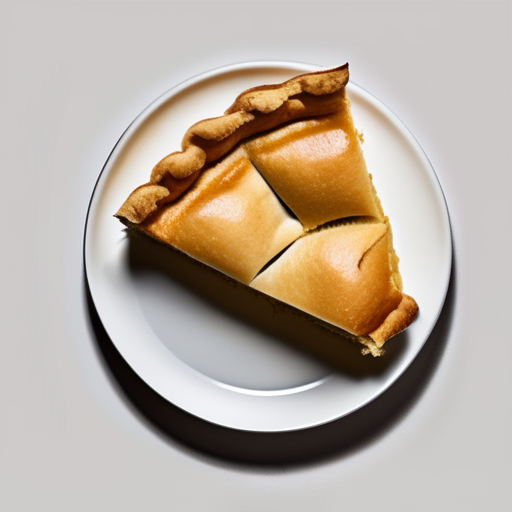}
  & \Figi{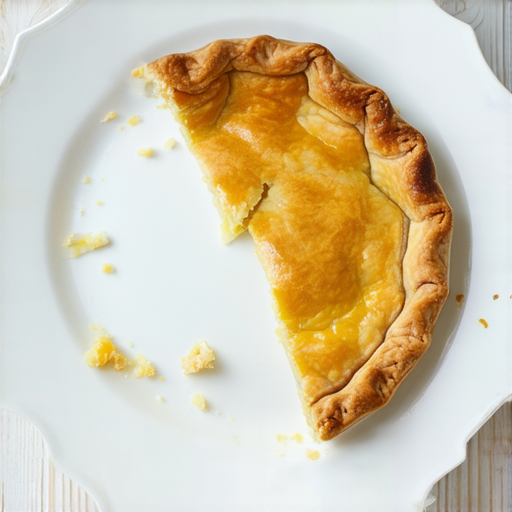}
  & \Figi{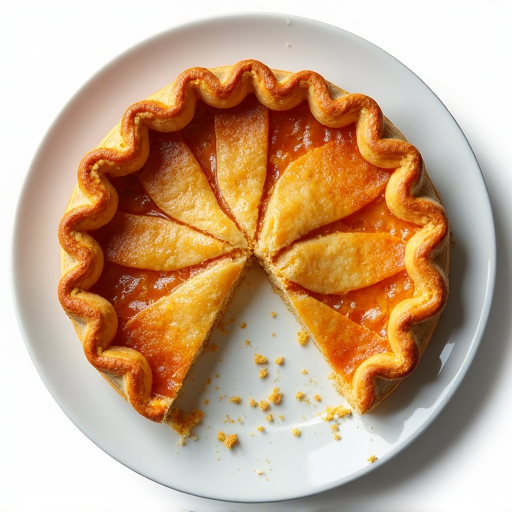}
  & \Figi{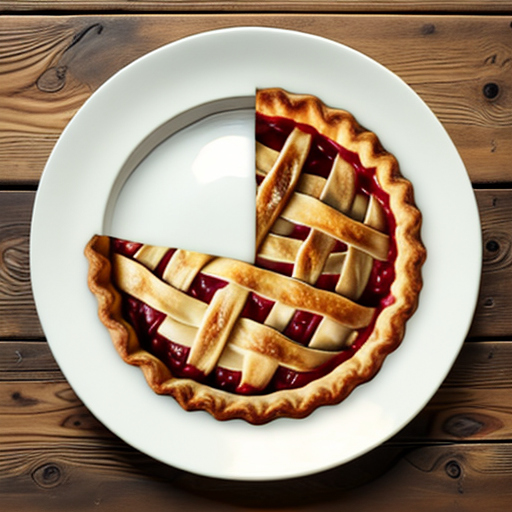} \\
\end{tabular}
\end{minipage}

\caption{
\textbf{Fraction generation.}
Base models frequently produce complete objects or conventional slices,
whereas our model accurately follows the requested fraction for both pizza
and pie across five target values.
}
\label{fig:fraction_baseline_qualitative}
\end{figure*}

\begin{table}[t]
\centering
\footnotesize
\setlength{\tabcolsep}{4pt}

\caption{Controlled comparison on a fixed SD 2.1 backbone
over the same 960 clock generations. Neither prompt
engineering nor text-conditioned fine-tuning provides
meaningful geometric control; only geometric tokens
achieve high accuracy.}

\label{tab:control_baselines}
\begin{tabular}{lccc}
\toprule
& \multicolumn{3}{c}{Acc@$6^\circ$ $\uparrow$} \\
\cmidrule(lr){2-4}
\textbf{Conditioning (SD~2.1)} & GPT-5.6 & Claude & IAT \\
\midrule
Text prompt                 & 0.4\% & 0.9\% & 0.5\% \\
~~+ prompt engineering      & 0.3\% & 0.7\% & 0.8\% \\
~~+ text-only LoRA          & 0.7\% & 0.2\% & 0.7\% \\
\midrule
~~+ geometric tokens (ours) & \textbf{71.8\%} & \textbf{63.1\%} & \textbf{77.7\%} \\
\bottomrule
\end{tabular}
\end{table}

\noindent\textbf{Quantitative results.}
Table~\ref{tab:clock_results} reports clock-generation performance
across automatic and human evaluators. 
Pretrained baselines achieve near-zero Acc@$6^\circ$,
whereas ours exceeds 70\% on unseen styles under GPT-5.6 (71.6\%) and IAT (79.4\%), and reaches 75.3\% under human annotation.
On 48 additional non-quarter-hour times,
our method achieves 77.6\% under IAT, closely matching its
77.7\% accuracy on quarter-hour targets. In the user study,
our method is preferred in 89.6\% of comparisons for
controllability and 50.3\% for appearance.

Table~\ref{tab:control_baselines} shows that, on the same
SD~2.1 backbone, 
prompt engineering achieves $0.3$--$0.8\%$ Acc@$6^\circ$
and text-only LoRA achieves $0.2$--$0.7\%$.
Thus, the gains stem from geometric tokens rather than
exposure to the training data.
Figure~\ref{fig:clock_comparison} illustrates the same failure mode:
baselines often generate plausible clocks whose hand configurations are
weakly related to the requested time, while our model adjusts both hands to
the target.

\begin{figure}[t]
\raggedright

\newlength{\Gcw}
\setlength{\Gcw}{1.62cm}

\newlength{\Glw}
\setlength{\Glw}{1.05cm}

\newcommand{\Gig}[1]{%
  \includegraphics[
    width=\Gcw,
    height=\Gcw,
    keepaspectratio=false
  ]{figs/gen/#1}%
}

\newcommand{\Glab}[1]{%
  \parbox[b][\Gcw][c]{\Glw}{\raggedleft\footnotesize #1}%
}

\newcommand{\Gcol}[1]{%
  \makebox[\Gcw][c]{\footnotesize\bfseries #1}%
}

\setlength{\tabcolsep}{0.5pt}
\renewcommand{\arraystretch}{0.88}

\begin{tabular}{
  @{}
  >{\arraybackslash}p{\Glw}
  @{\hspace{4pt}}
  cccc
  @{}
}
  &
  \Gcol{1/6} &
  \Gcol{1/4} &
  \Gcol{1/2} &
  \Gcol{3/4}
  \\[1pt]

  \Glab{Orange\\Slice}
    & \Gig{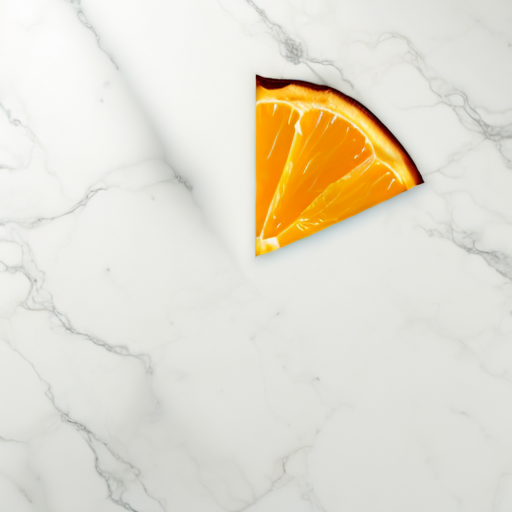}
    & \Gig{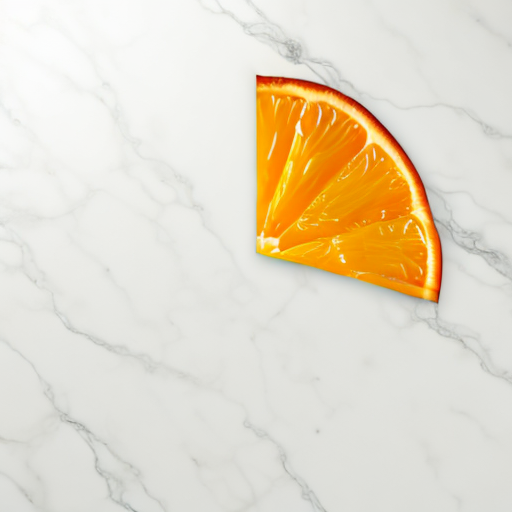}
    & \Gig{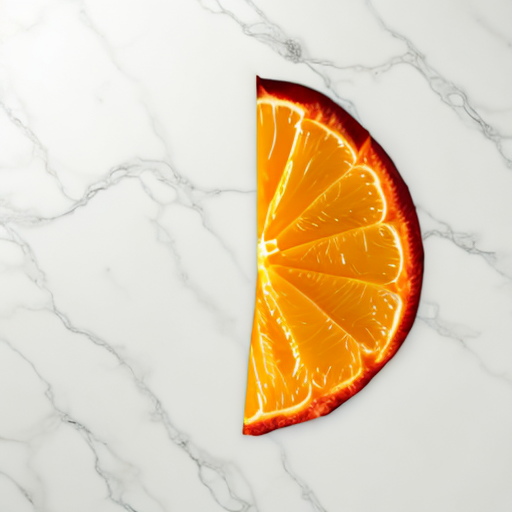}
    & \Gig{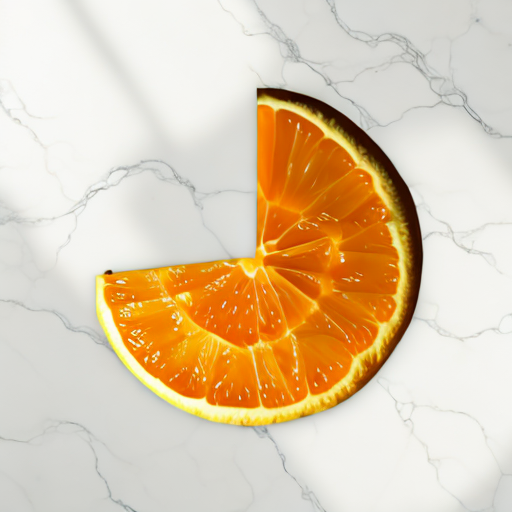}
    \\[-2pt]

  \Glab{Waffle}
    & \Gig{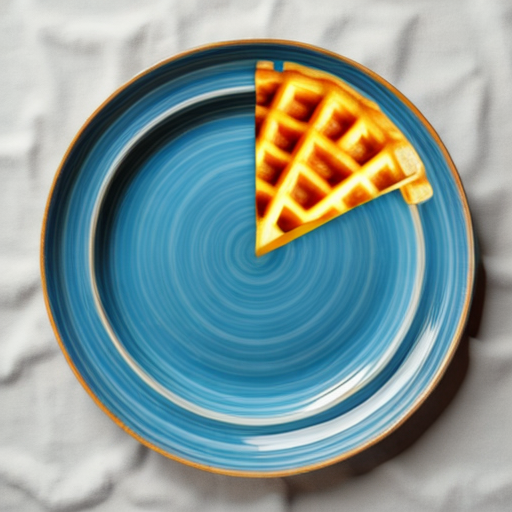}
    & \Gig{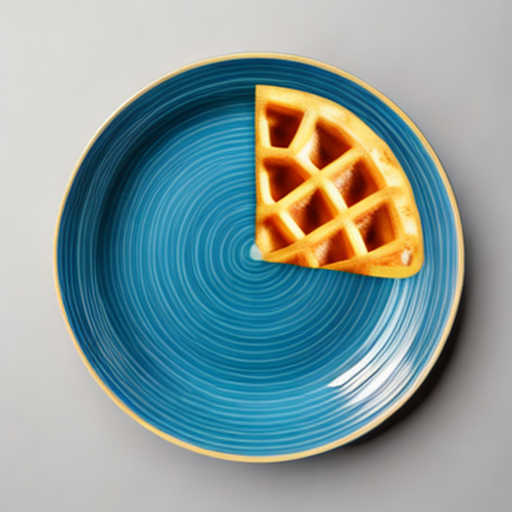}
    & \Gig{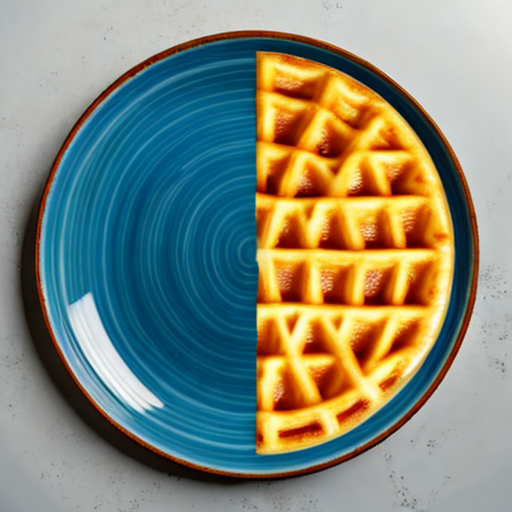}
    & \Gig{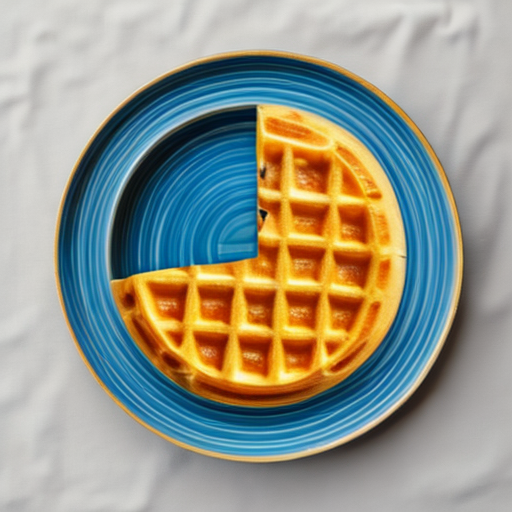}
    \\[-2pt]

  \Glab{Mosaic}
    & \Gig{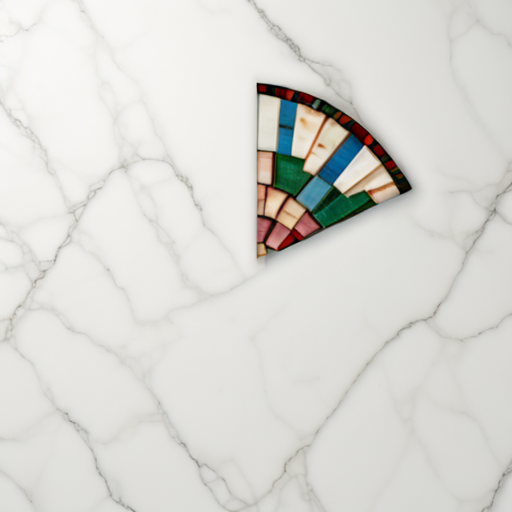}
    & \Gig{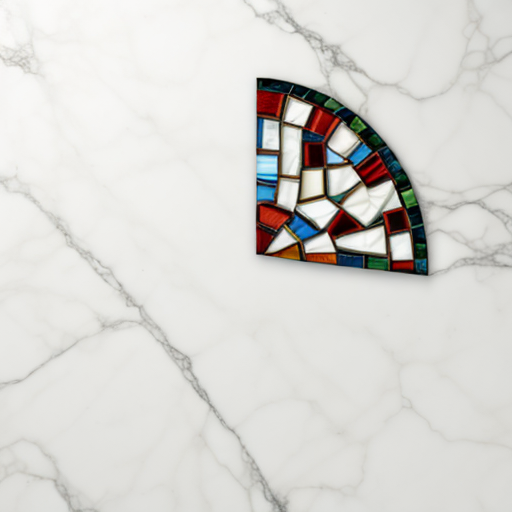}
    & \Gig{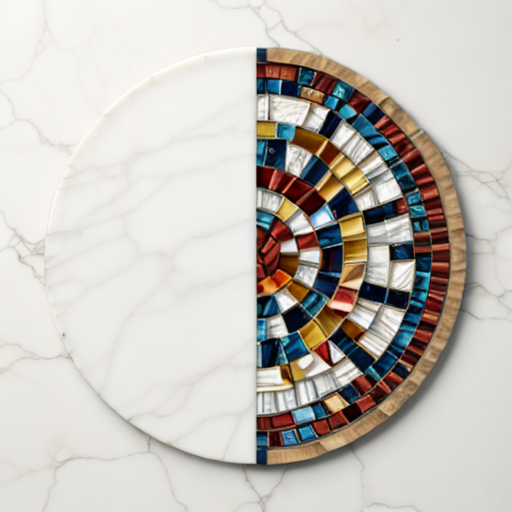}
    & \Gig{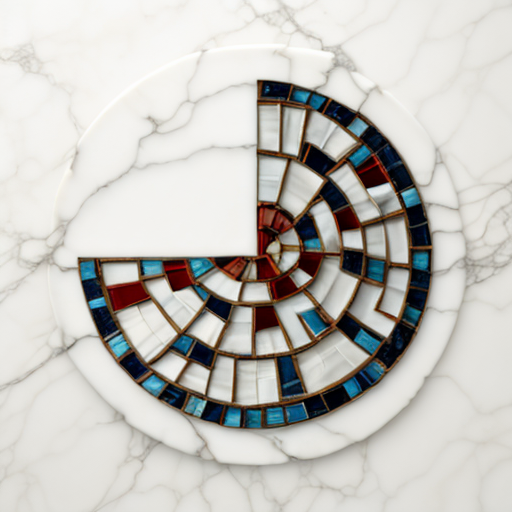}
    \\[-2pt]

  \Glab{Stained\\Glass}
    & \Gig{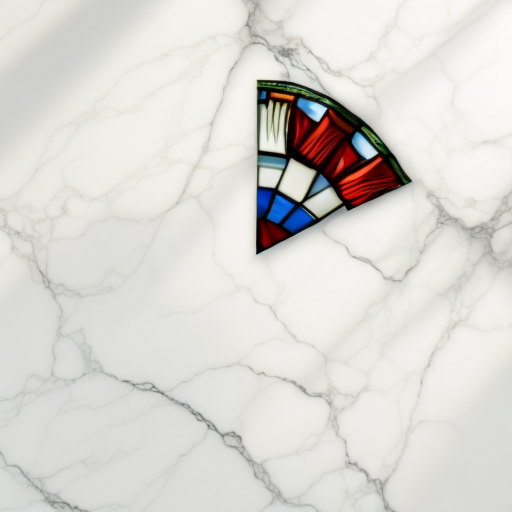}
    & \Gig{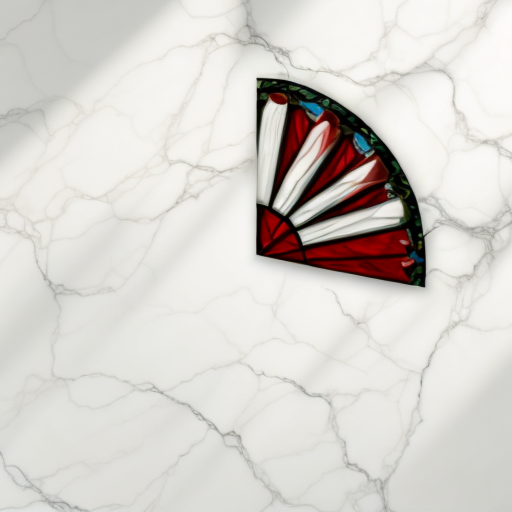}
    & \Gig{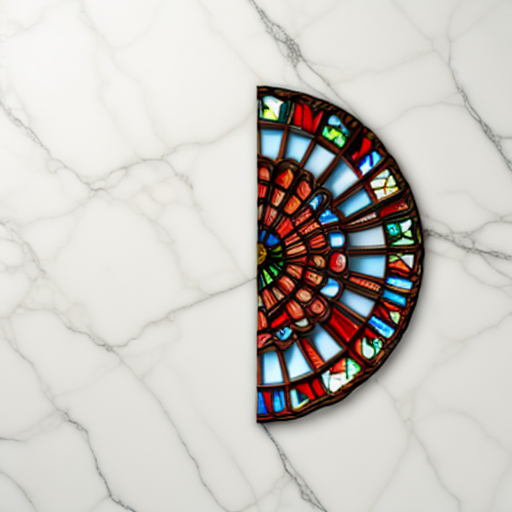}
    & \Gig{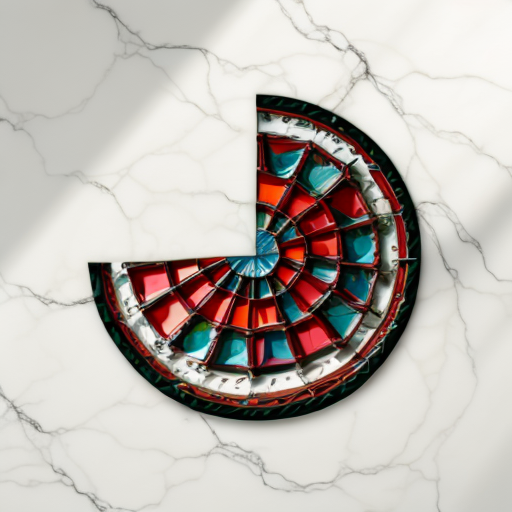}
\end{tabular}

\caption{%
  \textbf{Generalization to unseen fraction objects.}
  Our model follows the requested arc fraction across four object categories
  not seen during training (rows) and four target fractions (columns), without
  additional fine-tuning.
}
\label{fig:generalization}
\end{figure}

Table~\ref{tab:fraction_results} summarizes fraction controllability.
Pretrained baselines poorly match requested fractions, while
ours performs strongly on seen and unseen object categories.
On training objects, it reaches \textbf{89.5\% Acc@0.10}, and performance remains high for held-out fraction values, with \textbf{84.9\% Acc@0.10} on off-grid targets. A denser evaluation over all $1^\circ$ angular steps still yields \textbf{81.6\% Acc@0.10},
indicating that the model interpolates across the angular range rather than
memorizing the discrete values observed during training. In the user study, our method is selected in \textbf{92.8\%} of comparisons
for controllability and \textbf{67.4\%} for appearance quality. The qualitative comparison in Figure~\ref{fig:fraction_baseline_qualitative}
shows the same trend: base models frequently produce complete objects or
conventional slices, while our method adjusts the visible sector according
to the requested fraction while preserving a realistic appearance.

\noindent\textbf{Generalization to unseen appearances.}
We evaluate whether geometric control transfers to appearances not observed
during training. As shown in Figure~\ref{fig:generalization}, the fraction model
follows the requested angular extent on unseen objects including orange slices,
waffles, mosaics, and stained glass. The successful transfer to visually
distinct appearances such as mosaics and stained glass suggests that the learned
conditioning captures circular geometry rather than category-specific
appearance. As further illustrated in Figure~\ref{fig:teaser}, the model also
retains the requested geometric control when these objects are generated in
diverse scene contexts, with further in-scene results in Appendix~F.4.

\begin{table}[t]
\centering
\footnotesize
\setlength{\tabcolsep}{2pt}
\caption{Clock ablation (Acc@$6^\circ$: both hands within
$6^\circ$; averaged over train and unseen-style prompts).
Avg. MAE denotes the mean angular error across both hands.
Unless otherwise specified, ablations use the sinusoidal
encoder and SLR. IAT: ItsAboutTime ResNet50 reader.}
\label{tab:clock_ablation}
\begin{tabular}{lcccccc}
\toprule
& \multicolumn{2}{c}{\textbf{GPT-5.6}}
& \multicolumn{2}{c}{\textbf{Claude Opus 5}}
& \multicolumn{2}{c}{\textbf{IAT}} \\
\cmidrule(lr){2-3}
\cmidrule(lr){4-5}
\cmidrule(lr){6-7}

\textbf{Variant}
& A@$6^\circ$$\uparrow$ & MAE$\downarrow$
& A@$6^\circ$$\uparrow$ & MAE$\downarrow$
& A@$6^\circ$$\uparrow$ & MAE$\downarrow$ \\
\midrule

Ours
& \textbf{71.8\%} & 21.8$^\circ$
& 63.1\% & 24.2$^\circ$
& \textbf{77.7\%} & \textbf{18.8$^\circ$} \\

~~w/o swap pref.
& 66.8\% & 22.7$^\circ$
& 57.0\% & 26.5$^\circ$
& 65.6\% & 20.4$^\circ$ \\

\midrule

~~w/o SLR
& 20.3\% & 69.5$^\circ$
& 16.8\% & 73.1$^\circ$
& 23.5\% & 71.1$^\circ$ \\

~~w/o attn. local.
& 9.4\% & 53.5$^\circ$
& 8.9\% & 51.5$^\circ$
& 10.2\% & 50.1$^\circ$ \\

~~post-enc.\ injection
& 0.0\% & 97.3$^\circ$
& 0.0\% & 90.2$^\circ$
& 0.5\% & 88.3$^\circ$ \\

\midrule

~~raw scalar enc.
& 0.2\% & 85.1$^\circ$
& 0.0\% & 85.5$^\circ$
& 0.9\% & 84.5$^\circ$ \\

~~discrete bin enc.
& 42.9\% & 40.7$^\circ$
& 19.6\% & 41.7$^\circ$
& 54.9\% & 39.7$^\circ$ \\

\midrule
\multicolumn{7}{l}{
\textit{SLR mixture (component / object / unconstrained)}
} \\

~~55 / 30 / 15 (ours)
& \textbf{71.8\%} & 21.8$^\circ$
& 63.1\% & 24.2$^\circ$
& \textbf{77.7\%} & \textbf{18.8$^\circ$} \\

~~70 / 20 / 10
& 69.9\% & 23.7$^\circ$
& \textbf{63.7\%} & 26.6$^\circ$
& 72.5\% & 20.4$^\circ$ \\

~~40 / 40 / 20
& 57.4\% & 33.8$^\circ$
& 51.5\% & 37.8$^\circ$
& 63.4\% & 25.2$^\circ$ \\

~~33 / 33 / 33
& 66.7\% & \textbf{19.7$^\circ$}
& 58.3\% & \textbf{23.7$^\circ$}
& 64.8\% & 18.9$^\circ$ \\

~~20 / 40 / 40
& 57.7\% & 33.2$^\circ$
& 48.4\% & 36.7$^\circ$
& 62.4\% & 26.0$^\circ$ \\

\bottomrule
\end{tabular}
\end{table}

\noindent\textbf{Ablation study.}
We ablate the swap-preference objective, encoder representation, SLR,
attention localization, and geometric-token injection point. Clock ablations
use 20 style prompts over all 48 quarter-hour targets ($n=960$), covering
training and held-out styles, and are evaluated with GPT-5.6, Claude Opus~5,
and IAT.

Table~\ref{tab:clock_ablation} shows that the swap-preference loss improves
Acc@$6^\circ$ by 5.0--12.1 percentage points across evaluators. Removing SLR
reduces GPT-5.6 Acc@$6^\circ$ to $20.3$\%, while removing attention localization
further drops it to $9.4$\%, highlighting the importance of spatial grounding.
Post-encoder injection and raw-scalar encoding nearly eliminate controllability
($0.0$\% and $0.2$\%, respectively), whereas discrete binning recovers only
partial performance. Finally, among the tested SLR sampling mixtures, the
55/30/15 configuration achieves the highest Acc@$6^\circ$
under GPT-5.6 and IAT (71.8\% and 77.7\%, respectively).



\section{Conclusion}

We introduced a geometric token encoding framework for precise continuous angular control in text-to-image generation. By representing circular quantities with periodic role-specific tokens, using Stochastic Localization Relaxation (SLR) for training-time grounding, and introducing a clock-specific hand-swap preference objective, our method improves controllability for analog clocks and circular fractions. The learned control generalizes to unseen visual appearances and continuous target values.

\noindent\textbf{Limitations.}
Our method targets quantities with explicit circular geometric representation and requires task-specific training data with accurate geometric annotations. Controllability can degrade in visually complex scenes or when controlled components are ambiguous or poorly localized. Extending the framework to other geometric quantities
and reducing supervision remain promising directions.

{
    \small
    \bibliographystyle{ieeenat_fullname}
    \bibliography{main}
}

\newpage
\appendix

\section{Implementation Details}
\label{sec:implementation_details}

We use Stable Diffusion~2.1 as the backbone and train at a resolution of
$512\times512$. The text encoder remains frozen throughout training, while
LoRA adapters in the U-Net and the projection network $P$ are optimized.
We use LoRA rank 4, a batch size of 2 per GPU on two GPUs, and gradient
accumulation of 1. The base geometry model is trained for 20K steps using a
learning rate of $1\times10^{-4}$ for both the U-Net LoRA parameters and the
projection network.

For SLR, the component-level, object-level, and unconstrained regimes are
sampled with probabilities of 0.55, 0.30, and 0.15, respectively. Images are
center-cropped during preprocessing without color jitter, and mixed-precision
training is disabled.

For the final hand-swap preference refinement, we initialize from the 20K-step
geometry checkpoint and train for an additional 25K steps. During this stage,
the U-Net LoRA learning rate is reduced to $3\times10^{-5}$, while the
projection-network learning rate remains $1\times10^{-4}$. We set the
preference-loss weight $\lambda_{\mathrm{swap}}$ to 0.10 and $\beta$ to 5.0.
Inference is fully bounding-box-free and requires only the text prompt and
target quantity.

\subsection{Clock Training Data}
\label{sec:supp_clock_training}

Our clock training set contains 10,406 images, each derived
from a distinct procedurally rendered clock in the
\texttt{moondream/1M-synthetic-analog-clocks} dataset.
We restyle the source clocks using Stable Diffusion 3.5
Canny ControlNet and retain images without visible second
hands or logos, followed by visual quality filtering.
Stable Diffusion 3.5 is used only for data generation;
our trained backbone is Stable Diffusion 2.1.

Training times are inherited from the source dataset rather
than sampled from a fixed grid. The angular labels retain
second-level precision, and the training set covers 712
of the 720 possible minute positions on a 12-hour clock.
Quarter-hour configurations account for 7.8\% of the
images, while 20.1\% occur at five-minute intervals.
The canonical 10:10 configuration is not overrepresented:
its local frequency is approximately 6.1 images per
minute position, compared with a dataset-wide average
of 14.5.

Both training stages use the same image set. The initial
geometry training uses a single geometric prompt template,
while the preference-refinement stage uses the 12 training
style templates described in
Sec.~\ref{sec:supp_clock_protocol}.

\section{User Study and Human Annotation}
\label{sec:user_study}

\subsection{User Study}

We conducted a human preference study to evaluate two complementary aspects
of the generated images: \emph{controllability} and \emph{appearance quality}.
A total of 33 participants completed the study, which included 25 analog-clock
and 25 fraction examples per participant.

For each example, participants were shown the text prompt together with five
anonymized images generated by our method and four baselines: Stable
Diffusion~2.1, SDXL, Stable Diffusion~3.5, and Flux.1. Model identities were
hidden, and images were labeled using neutral identifiers (Image A--E).
Image positions were randomized and balanced to reduce ordering and position
bias. The prompt explicitly included the target time for clocks or the target
fraction.

Participants answered two questions for each example. For controllability,
they selected the image that best represented the specified target time or
fraction. For appearance quality, they selected the image that best matched
the object, material, setting, and other visual attributes described in the
prompt, independently of target-geometry accuracy. This yields 825 selections
per task and criterion.

Overall, our method was selected in 91.2\% of controllability evaluations and
58.8\% of appearance evaluations. For clocks, it received 89.6\% and 50.3\%,
respectively; for fractions, 92.8\% and 67.4\%.

\begin{figure*}[t]
    \centering
    \begin{minipage}[t]{0.49\linewidth}
        \centering
        \includegraphics[width=\linewidth]{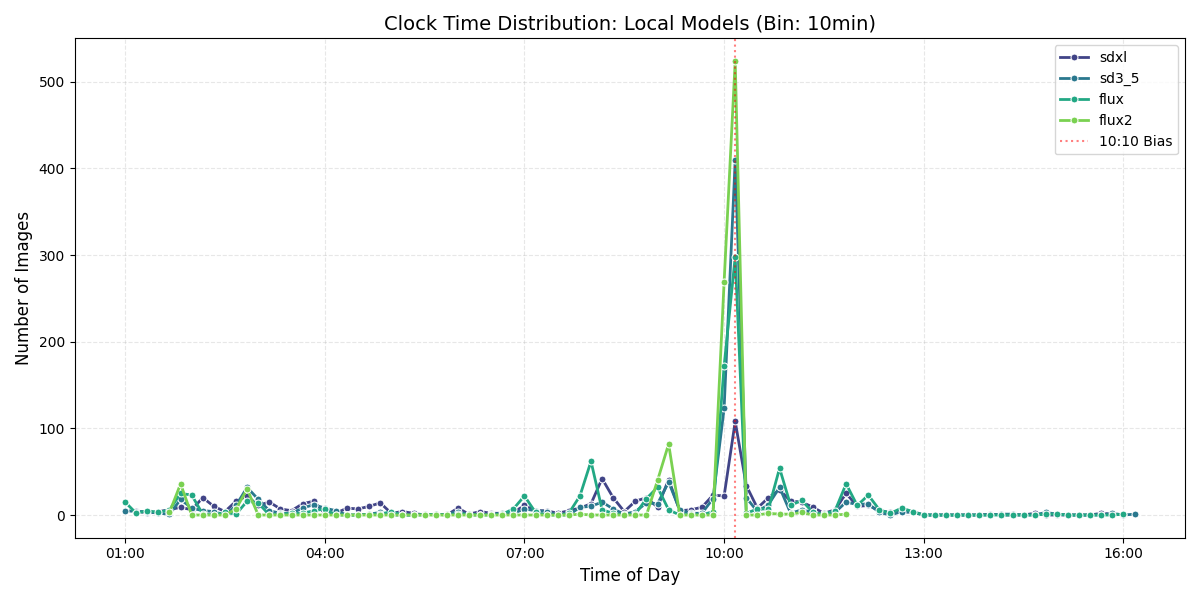}
        \vspace{-2mm}
        \small (a) Local models
    \end{minipage}
    \hfill
    \begin{minipage}[t]{0.49\linewidth}
        \centering
        \includegraphics[width=\linewidth]{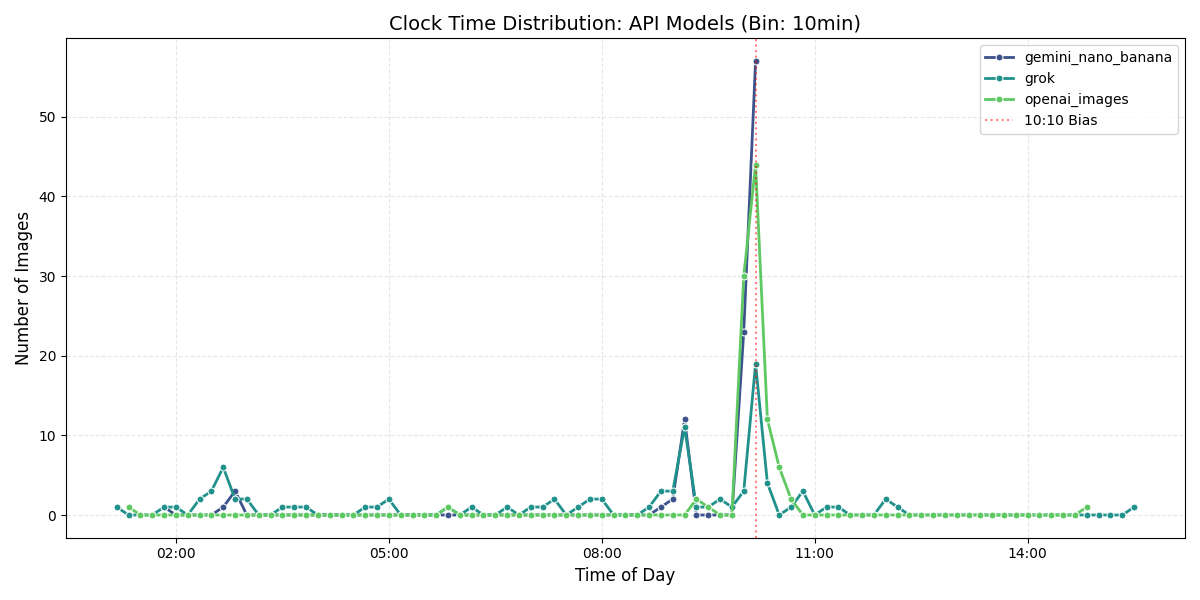}
        \vspace{-2mm}
        \small (b) API models
    \end{minipage}
    \caption{
    Distribution of generated analog-clock times for pretrained text-to-image
    models. Several models exhibit strong modes near canonical configurations,
    particularly 10:10, despite variation in the requested target time.
    }
    \label{fig:clock_time_bias}
\end{figure*}

\subsection{Human Clock Annotation}
\label{sec:supp_human_annotation}

In addition to the preference-based user study, we
conducted a separate human clock-reading evaluation.
Three annotators independently read 140 generated clock
images, yielding 420 annotations in total. The evaluation
set includes 100 images from our method, approximately
balanced across all 20 clock styles, and 40 baseline
images, 10 from each.

The target times are selected from the 48 quarter-hour
configurations used in the main evaluation. Annotators
provide the displayed time in \texttt{HH:MM} format
or mark the clock as unreadable (\texttt{U}).
Unreadable responses are counted as incorrect.
The response rate is approximately 98\% for our
method, compared with 33\%--80\% for the baselines.

\section{Canonical Clock-Time Bias}
\label{sec:clock_bias}

To characterize the failure mode of pretrained text-to-image models, we analyze
the distribution of clock times present in their generated images.
Figure~\ref{fig:clock_time_bias} shows the predicted-time distributions for
local and API-based generators, while Table~\ref{tab:clock_bias} summarizes
the dominant readable clock configuration for each model.

\begin{table}[t]
\centering
\footnotesize
\setlength{\tabcolsep}{4pt}
\caption{Canonical clock-time bias in pretrained text-to-image models.
We report the most frequent readable clock configuration and its frequency
among readable generations.}
\label{tab:clock_bias}
\begin{tabular}{lccc}
\toprule
\textbf{Model} & \textbf{N} & \textbf{Top time} &
\textbf{Frequency} \\
\midrule
SD~3.5             & 1000   & 10:10 & 33.7\% \\
Flux.1             & 1000 & 10:10 & 26.2\% \\
Flux.2             & 1000 & 10:10 & 48.0\% \\
Gemini Nano Banana & 100   & 10:10 & 47.0\% \\
\bottomrule
\end{tabular}
\end{table}

A pronounced concentration around common clock configurations is visible
across several models, particularly near 10:10. The canonical 10:10 pose is
the most frequent output for SD~3.5, Flux.1, Flux.2, and Gemini Nano Banana,
accounting for 33.7\%, 26.2\%, 48.0\%, and 47.0\% of their readable
generations, respectively. Grok also exhibits a noticeable mode near this
configuration, while other models show related concentrations in the
10-o'clock region.

Importantly, these modes emerge despite variation in the requested target
times. The failures are therefore not limited to small angular deviations from
the requested configuration; instead, pretrained generators often collapse
toward a small set of visually common clock poses. This behavior supports our
motivation for explicitly conditioning generation on the requested geometric
state rather than relying on natural-language specification alone.

\section{Prompt Templates and Generation Settings}
\label{sec:supp_prompts}

Every quantitative result in the paper is produced under one of two prompting
protocols. Baselines have no geometric conditioning, so the target quantity is
written into the prompt as text. Our method never receives the target as text:
the prompt carries a pair of geometric tokens, written here as
$\langle e_h\rangle,\langle e_m\rangle$ for the hour and minute hands and
$\langle e_s\rangle,\langle e_e\rangle$ for the start and end boundaries of a
fraction sector, whose embeddings the projection network $P$ produces from the sinusoidal
encoding of the target angles and substitutes before the first CLIP
transformer layer. Apart from that substitution, the two protocols share the
same sentence, the same style vocabulary, the same targets and the same seed
schedule, so a comparison isolates the conditioning mechanism.

\subsection{Clock Protocol}
\label{sec:supp_clock_protocol}

All clock results---the main comparison, architecture ablations, and SLR
mixture study---use the same evaluation grid: 20 style prompts $\times$ 48
quarter-hour targets, giving $n=960$ images per model. The first 12 styles
appear in training; the remaining 8 are held out for unseen-style evaluation.

To evaluate finer-grained temporal control, we additionally
sample 48 non-quarter-hour target times, selecting four
times from each of the 12 hours with minute values that
are not multiples of five. We generate each target using
the same 20 style prompts, yielding 960 additional images.
These targets are distinct from the quarter-hour evaluation
grid but are not held out from training, since the training
set covers nearly all minute positions.

Each style is specified by a template containing a single
\texttt{PLACEHOLDER}; Table~\ref{tab:supp_styles} lists all 20. For baselines,
the placeholder is replaced by \texttt{a clock showing HH:MM}; for our method,
it is replaced by \texttt{a $\langle e_h\rangle$ hour hand and a
$\langle e_m\rangle$ minute hand}. Nothing else in the template changes.

\begin{table}[h]
\centering
\footnotesize
\setlength{\tabcolsep}{4pt}
\caption{The 20 clock style templates. Styles 1--12  are training styles; 13--20 are held out and used for the unseen-style evaluation.}
\label{tab:supp_styles}
\begin{tabular}{@{}rp{0.88\linewidth}@{}}
\toprule
\# & \textbf{Template} \\
\midrule
\multicolumn{2}{@{}l}{\emph{Training styles}} \\
1  & PLACEHOLDER, a realistic analog wall clock \\
2  & PLACEHOLDER, a decorative analog clock \\
3  & PLACEHOLDER, a black analog clock \\
4  & PLACEHOLDER, a vintage analog clock \\
5  & PLACEHOLDER, a minimalist white clock \\
6  & PLACEHOLDER, a large analog clock \\
7  & PLACEHOLDER, a worn analog clock \\
8  & PLACEHOLDER, a metal analog clock \\
9  & PLACEHOLDER, a round analog clock \\
10 & PLACEHOLDER, a brass analog clock \\
11 & PLACEHOLDER, a colorful analog clock \\
12 & PLACEHOLDER, a small analog alarm clock \\
\midrule
\multicolumn{2}{@{}l}{\emph{Unseen styles}} \\
13 & PLACEHOLDER on a blue enamel industrial analog wall clock with an off-white face, workshop product photo \\
14 & PLACEHOLDER on an antique brass art deco table clock, dark navy dial, ornate metal frame, warm tungsten light \\
15 & PLACEHOLDER on a brushed stainless steel analog clock with crisp black numerals, realistic studio lighting \\
16 & PLACEHOLDER on a green marble luxury analog clock, gold bezel, polished stone texture, elegant hotel interior \\
17 & PLACEHOLDER on a matte black ceramic analog wall clock with a thin silver rim, product photography \\
18 & PLACEHOLDER on a glossy red plastic retro kitchen clock, rounded vintage case, white dial, colorful 1960s style \\
19 & PLACEHOLDER on a transparent acrylic minimal analog clock body, floating black ticks, clean white background \\
20 & PLACEHOLDER on a warm oak wood analog wall clock, Scandinavian design, cream dial, natural grain, soft daylight \\
\bottomrule
\end{tabular}
\end{table}

\paragraph{Example.}
For style 3 at target 07:15, the two protocols give
\begin{quote}\footnotesize
\textbf{Baselines:} a clock showing 07:15, a black analog clock \\
\textbf{Ours:} a $\langle e_h\rangle$ hour hand and a
$\langle e_m\rangle$ minute hand, a black analog clock
\end{quote}

\paragraph{Negative prompt.}
Baselines use the negative prompt
\texttt{wrong hands, distorted clock, unreadable clock, extra hands,
missing hands, blurry, low quality}, which provides an additional prompting
advantage to the baselines. Our model uses no negative prompt.

\subsection{Clock Reader Evaluation}
\label{sec:supp_clock_readers}

We evaluate clock readability using GPT-5.6
(\texttt{gpt-5.6-sol}), Claude Opus 5
(\texttt{claude-opus-5}), and ItsAboutTime (IAT).
Each vision-language model receives one request
per image with the identical prompt:

\begin{quote}
\small
Look at this analog clock image carefully.
What time is displayed on the clock?
Reply with ONLY the time in HH:MM format
(24h or 12h, e.g. 06:00 or 14:30).
No explanation, just the time.
\end{quote}

We extract the first valid \texttt{HH:MM} prediction
from each response. Missing or unparsable responses
are counted as incorrect. The maximum output token
budgets are 512 for GPT-5.6 and 1024 for Claude Opus 5.

For IAT, we use the \texttt{full+++} checkpoint,
consisting of a spatial transformer network (STN)
followed by a ResNet50-based classifier. The reader
operates on $224\times224$ images and predicts one
of 720 minute-resolution clock configurations.

\subsection{Fraction Protocol}
\label{sec:supp_fraction_protocol}

Fraction results use 71 targets per object, sweeping from $5^\circ$ to
$355^\circ$ in $5^\circ$ steps with the start fixed at 12 o'clock.
Object templates are photorealistic descriptions containing a single
\texttt{PLACEHOLDER} slot; the splits differ only in the instantiated object
templates.

For baselines, the slot becomes an explicit textual fraction:
\texttt{showing only $P$\% of its area as a single arc sector starting at
12 o'clock and going clockwise, the rest is absent}. For our method, it becomes
\texttt{with a $\langle e_s\rangle$ start boundary and a
$\langle e_e\rangle$ end boundary}. Thus, our prompt is constant across all
71 targets for a given object; only the angles provided to the geometric
encoder change.

\paragraph{Example.}
For the pepperoni pizza template at $f=0.25$:
\begin{quote}\footnotesize
\textbf{Baselines:} a top-down photo of a round pepperoni pizza with evenly
spread pepperoni slices and bubbling cheese showing only 25\% of its area as a
single arc sector starting at 12 o'clock and going clockwise, the rest is
absent, on a white marble kitchen countertop, soft daylight

\textbf{Ours:} a top-down photo of a round pepperoni pizza with evenly spread
pepperoni slices and bubbling cheese with a $\langle e_s\rangle$ start boundary
and a $\langle e_e\rangle$ end boundary, on a white marble kitchen countertop,
soft daylight
\end{quote}

\subsection{Fraction Evaluation}
\label{sec:supp_fraction_eval}

We evaluate fraction generation using a training-free,
rule-based angular segmentation detector. Assuming
the object is approximately centered, the detector
estimates its radius and extracts angular color
profiles from an annular region spanning 12\%--68\%
of the estimated radius. At 360 angular positions measured clockwise from
12 o'clock, we compute mean RGB, saturation, and
brightness features. The predicted fraction boundary
minimizes the sum of within-region squared errors
for the filled and empty sectors. On 1,000 validation images with known sector angles,
the detector achieves 91.7\% Acc@0.10, increasing
to 99.9\% for targets between $45^\circ$ and
$315^\circ$.

\subsection{Prompt-Engineering Baseline}
\label{sec:supp_prompt_eng}

To test whether careful prompting alone can recover angular control, we repeat
the clock grid with the target expressed as explicit hand geometry rather than
as a clock reading. The carrier phrase specifies the hand orientations
corresponding to each quarter-hour target:

\begin{quote}\footnotesize
\texttt{:00} $\rightarrow$ the hour hand pointing at $h$ and the minute hand pointing at 12 \\
\texttt{:15} $\rightarrow$ the hour hand pointing just past $h$ and the minute hand pointing at 3 \\
\texttt{:30} $\rightarrow$ the hour hand pointing halfway between $h$ and $h{+}1$ and the minute hand pointing at 6 \\
\texttt{:45} $\rightarrow$ the hour hand pointing just before $h{+}1$ and the minute hand pointing at 9
\end{quote}

For style 3 at 01:15, this yields
\texttt{a clock with the hour hand pointing just past 1 and the minute hand
pointing at 3, a black analog clock}. Styles, targets, seeds, sampler settings,
and negative prompts are otherwise identical to
Sec.~\ref{sec:supp_clock_protocol}, so the runs differ only in conditioning
phrasing.

\subsection{Sampling Settings}
\label{sec:supp_sampling}

Sampling settings are summarized in Table~\ref{tab:supp_sampling}. We use the
standard sampling configuration associated with each pretrained baseline,
while our model follows the Stable Diffusion~2.1 setting. All images are
generated at $512\times512$ resolution.

\begin{table}[h]
\centering
\footnotesize
\setlength{\tabcolsep}{5pt}
\caption{Sampler configuration. Each model uses its corresponding step count
and guidance scale; all main-table images are generated at
$512{\times}512$.}
\label{tab:supp_sampling}
\begin{tabular}{@{}lrrr@{}}
\toprule
\textbf{Model} & \textbf{Steps} & \textbf{Guidance} & \textbf{Resolution} \\
\midrule
SD 2.1  & 50 & 7.5 & $512^2$ \\
SDXL    & 50 & 7.5 & $512^2$ \\
SD 3.5  & 28 & 4.5 & $512^2$ \\
Flux.1  & 28 & 3.5 & $512^2$ \\
Ours    & 50 & 7.5 & $512^2$ \\
\bottomrule
\end{tabular}
\end{table}

\section{Teaser Figure Prompts}
\label{sec:supp_teaser_prompts}

This section gives the complete prompt specification for the teaser figure
(Figure~1). The two protocols shown in the figure differ only in
how the geometric quantity reaches the model.

\paragraph{Protocol.}
Baselines receive no geometric conditioning, so the target is written into the
prompt as text: a clock reading (\eg ``showing 7:15'') for the clock domain and
a percentage of area (\eg ``showing only 40\% of its area'') for the fraction
domain. Our method never sees the target as text. The prompt instead carries a
pair of geometric tokens whose embeddings are produced by the projection network $P$ from
the sinusoidal encoding of the target angles and substituted before the first
CLIP transformer layer. We write these tokens as $\langle e_h\rangle$ and
$\langle e_m\rangle$ for the hour and minute hands, and as $\langle e_s\rangle$
and $\langle e_e\rangle$ for the start and end boundaries of a fraction
sector. The remainder of each prompt --- object, surrounding scene, lighting
phrase --- is held identical across protocols, so a column differs only in how
the angle is supplied.

\subsection{Clock Columns}

The geometric token pair is not tied to a fixed position in the prompt. Two of
the four columns place it at the front, before the object description; the
other two place it mid-sentence, after the object and scene have been named.
Both instantiations are shown, and both hit their target, which illustrates
that the tokens act through the pre-encoder substitution rather than through a
particular surface word order.

\begin{table}[h]
\centering
\footnotesize
\setlength{\tabcolsep}{4pt}
\caption{Clock prompts. $\langle e_h\rangle$ and $\langle e_m\rangle$ carry the
target hand angles; no textual time appears in our prompts.}
\label{tab:supp_teaser_clock}
\begin{tabular}{@{}p{0.10\linewidth}p{0.12\linewidth}p{0.74\linewidth}@{}}
\toprule
\textbf{Target} & \textbf{Row} & \textbf{Prompt} \\
\midrule
7:15 & Baselines & an analog clock showing 7:15, a realistic analog wall clock in a bright kitchen with coffee mugs and fruit on the counter, high detail, natural lighting \\
     & Ours      & a $\langle e_h\rangle$ hour hand and a $\langle e_m\rangle$ minute hand, a realistic analog wall clock in a bright kitchen with coffee mugs and fruit on the counter, high detail, natural lighting \\
\midrule
3:00 & Baselines & an analog clock showing 3:00, a large analog clock on a cafe wall above small tables and chairs, high detail, natural lighting \\
     & Ours      & a $\langle e_h\rangle$ hour hand and a $\langle e_m\rangle$ minute hand, a large analog clock on a cafe wall above small tables and chairs, high detail, natural lighting \\
\midrule
7:00 & Baselines & a round analog clock in a greenhouse, showing 7:00, with plants and clay pots nearby, high detail, natural lighting \\
     & Ours      & a round analog clock in a greenhouse, showing a $\langle e_h\rangle$ hour hand and a $\langle e_m\rangle$ minute hand, with plants and clay pots nearby, high detail, natural lighting \\
\midrule
11:45 & Baselines & an elegant brass analog clock in a hotel lobby, showing 11:45, with plants and a sofa nearby, high detail, natural lighting \\
      & Ours      & an elegant brass analog clock in a hotel lobby, showing a $\langle e_h\rangle$ hour hand and a $\langle e_m\rangle$ minute hand, with plants and a sofa nearby, high detail, natural lighting \\
\bottomrule
\end{tabular}
\end{table}

\subsection{Fraction Columns}

For our method the prompt is \emph{identical} across all four fraction columns;
only the angles fed to the encoder change. This is the point of the comparison:
one sentence yields four different filled sectors.

\begin{table}[h]
\centering
\footnotesize
\setlength{\tabcolsep}{4pt}
\caption{Fraction prompts. $\langle e_s\rangle$ and $\langle e_e\rangle$ encode
the start and end boundaries of the sector; the fraction never appears as text.}
\label{tab:supp_teaser_fraction}
\begin{tabular}{@{}p{0.10\linewidth}p{0.12\linewidth}p{0.75\linewidth}@{}}
\toprule
\textbf{Target} & \textbf{Row} & \textbf{Prompt} \\
\midrule
2/5 & Baselines & a top-down realistic food scene with a giant round chocolate chip cookie showing only 40\% of its area as a filled arc sector starting at 12 o'clock going clockwise, next to a glass of milk and chocolate pieces, on a kitchen counter \\
    & Ours      & a top-down realistic food scene with a giant round chocolate chip cookie with a $\langle e_s\rangle$ start boundary and a $\langle e_e\rangle$ end boundary, next to a glass of milk and chocolate pieces, on a kitchen counter \\
\midrule
4/5 & Baselines & a top-down realistic food scene with a round pumpkin pie showing only 80\% of its area as a filled arc sector starting at 12 o'clock going clockwise, next to cinnamon sticks and a tea cup, on a warm wooden table \\
    & Ours      & a top-down realistic food scene with a round pumpkin pie with a $\langle e_s\rangle$ start boundary and a $\langle e_e\rangle$ end boundary, next to cinnamon sticks and a tea cup, on a warm wooden table \\
\midrule
3/4 & Baselines & a top-down realistic food scene with a round red velvet cake showing only 75\% of its area as a filled arc sector starting at 12 o'clock going clockwise, next to a cup of tea and berries, on a terrazzo table \\
    & Ours      & a top-down realistic food scene with a round red velvet cake with a $\langle e_s\rangle$ start boundary and a $\langle e_e\rangle$ end boundary, next to a cup of tea and berries, on a terrazzo table \\
\midrule
1/4 & Baselines & a top-down realistic food scene with a round margherita pizza showing only 25\% of its area as a filled arc sector starting at 12 o'clock going clockwise, next to a small bowl of olives and a lemonade glass, on a light oak table \\
    & Ours      & a top-down realistic food scene with a round margherita pizza with a $\langle e_s\rangle$ start boundary and a $\langle e_e\rangle$ end boundary, next to a small bowl of olives and a lemonade glass, on a light oak table \\
\bottomrule
\end{tabular}
\end{table}

\section{Additional Results}

\subsection{SLR Mixture Selection and Seed Variability}
\label{sec:supp_slr_validation}

To independently evaluate the chosen SLR mixture, we
construct a disjoint validation split containing 48 non-quarter-hour clock times and 20 style prompts (12 training styles and 8 validation-only styles).
The target times, unseen styles, and random seeds do not
overlap with either test evaluation set.

We evaluate five SLR mixtures using the IAT clock reader.
Table~\ref{tab:slr_validation} reports validation and
test Acc@$6^\circ$ for a fixed training seed (1908).
The 55/30/15 mixture achieves the highest validation
accuracy (78.6\%), followed by 70/20/10 (76.2\%).
A paired bootstrap comparison yields a validation
difference of 2.4 percentage points with a 95\%
confidence interval of [-1.0, 5.6], indicating that
the single-seed difference is not statistically
conclusive.

\begin{table}[t]
\centering
\footnotesize
\setlength{\tabcolsep}{5pt}
\caption{SLR mixture comparison using IAT Acc@$6^\circ$.
Mixtures are ordered as component/object/unconstrained.
All results use training seed 1908.}
\label{tab:slr_validation}
\begin{tabular}{lcc}
\toprule
\textbf{Mixture} &
\textbf{Validation} &
\textbf{Test} \\
\midrule
55 / 30 / 15 & \textbf{78.6\%} & \textbf{77.7\%} \\
70 / 20 / 10 & 76.2\% & 72.5\% \\
40 / 40 / 20 & 73.9\% & 63.4\% \\
33 / 33 / 33 & 72.9\% & 64.8\% \\
20 / 40 / 40 & 70.4\% & 62.4\% \\
\bottomrule
\end{tabular}
\end{table}

To assess training-seed sensitivity, we additionally
retrain the two strongest validation mixtures with
seeds 2027 and 3141, for three independent training
runs per mixture. Table~\ref{tab:slr_seeds} reports
the mean and standard deviation across training seeds.
The 55/30/15 mixture outperforms 70/20/10 on both
validation and test sets for every training seed.
However, both configurations exhibit substantial
seed-dependent variability, particularly in the
frequency of hour--minute hand swaps.

\begin{table}[t]
\centering
\footnotesize
\setlength{\tabcolsep}{4pt}
\caption{Training-seed variability across three runs
(seeds 1908, 2027, and 3141), evaluated with IAT.
Values are mean $\pm$ standard deviation of
Acc@$6^\circ$ across training seeds.}
\label{tab:slr_seeds}
\begin{tabular}{lcc}
\toprule
\textbf{Mixture} &
\textbf{Validation} &
\textbf{Test} \\
\midrule
55 / 30 / 15 &
$\mathbf{74.4 \pm 11.0}$ &
$\mathbf{72.4 \pm 10.7}$ \\
70 / 20 / 10 &
$65.1 \pm 14.1$ &
$63.8 \pm 9.7$ \\
\bottomrule
\end{tabular}
\end{table}

These results support selecting 55/30/15 based on
validation performance rather than test-set tuning.
While its advantage is consistent across the three
tested training seeds, the substantial seed variability
indicates that improving hand-role assignment
robustness remains an important direction.

\begin{table}[t]
\centering
\footnotesize
\caption{
Fraction ablation on $n=994$ samples
(14 training object categories, 71 angles).
Acc@$k$: fraction error $\leq k$.}
\label{tab:fraction_ablation}
\setlength{\tabcolsep}{3pt}
\begin{tabular}{l r c c c c}
\toprule
\textbf{Variant} & $n$ & \textbf{MAE}$\downarrow$
& \textbf{Acc@0.05}$\uparrow$
& \textbf{Acc@0.10}$\uparrow$
& \textbf{Spear.}$\uparrow$ \\
\midrule
Ours
& 994 & \textbf{.065} & 82.0\% & \textbf{89.5\%} & \textbf{.821} \\
\midrule
w/o SLR
& 994 & .109 & 69.1\% & 80.7\% & .661 \\
w/o attn.\ localization
& 994 & .068 & \textbf{82.1\%} & 88.5\% & .790 \\
Post-enc.\ inj.
& 994 & .145 & 56.3\% & 69.3\% & .548 \\
\midrule
Raw scalar
& 994 & .083 & 71.4\% & 85.7\% & .765 \\
Discrete bins
& 994 & .116 & 65.1\% & 72.8\% & .639 \\
\bottomrule
\end{tabular}
\end{table}

\begin{figure*}[t]
    \centering
    \includegraphics[width=0.88\linewidth]{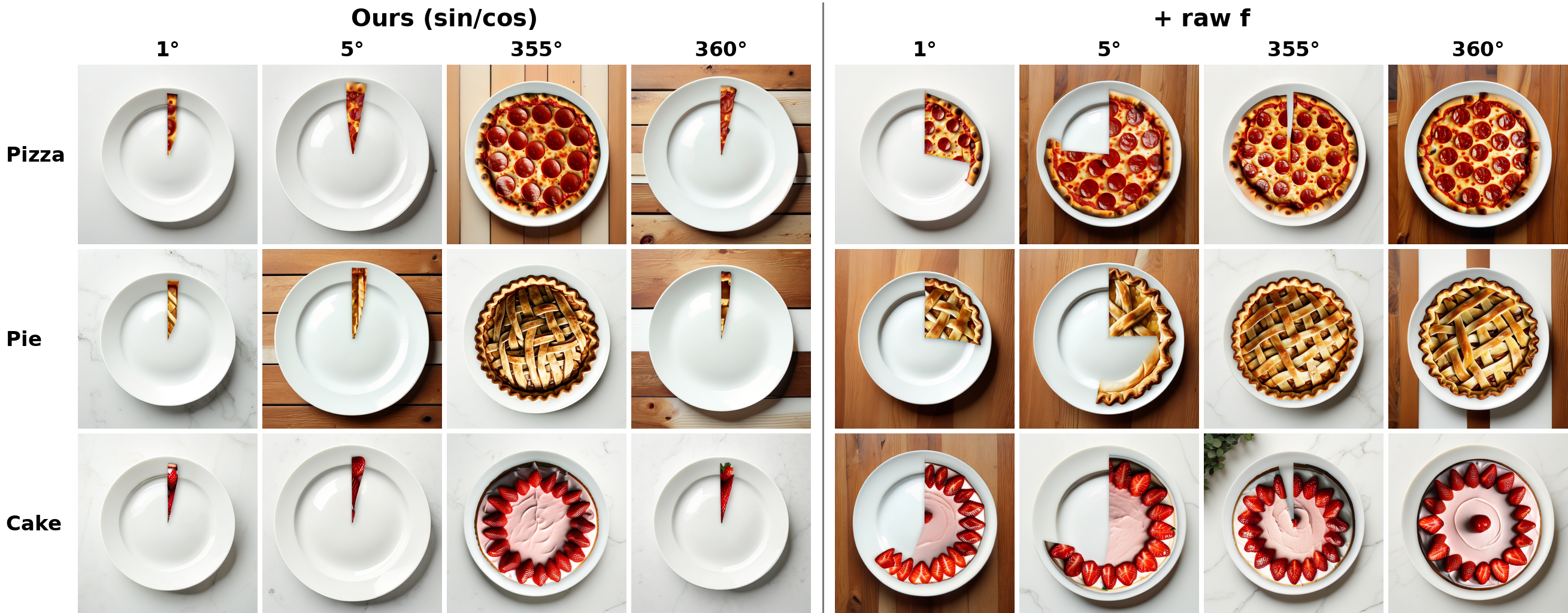}
    \caption{
    Qualitative comparison of the original periodic encoding
    and the augmented encoding with the scalar fraction $f$
    near the angular boundaries. The periodic encoding
    produces small sectors near $0^\circ$ but may generate
    narrow slices even for complete-sector targets
    ($360^\circ$), reflecting its endpoint ambiguity.
    Adding $f$ distinguishes the endpoints and enables
    complete-object generation, but can produce oversized
    sectors for small target angles ($1^\circ$--$5^\circ$).
    Results are shown for pizza, pie, and cake.
    }
    \label{fig:fraction_boundary}
\end{figure*}

\subsection{Fraction Ablations}

Table~\ref{tab:fraction_ablation} reports ablations for the fraction model.
Removing SLR reduces Acc@0.10 from \textbf{89.5\%} to $80.7\%$,
showing that exposure to weaker spatial supervision improves
bounding-box-free inference. Removing attention localization has a much smaller effect
($88.5\%$), suggesting that the fraction task is less sensitive to explicit
spatial grounding than clock generation. Post-encoder injection and discrete
binning substantially reduce performance, while raw scalar encoding remains
competitive but below the periodic sin--cos representation.

\subsection{Boundary Behavior of Fraction Encoding}
\label{sec:fraction_boundary}

Our periodic fraction encoding uses a fixed start angle
$\theta_s=0$ and an end angle $\theta_e=2\pi f$.
Consequently, $f=0$ and $f=1$ produce identical
representations, making empty and complete sectors
indistinguishable from the encoding alone.

We explored an augmented encoding that additionally
includes the scalar fraction value,
$[f,\sin(2\pi f),\cos(2\pi f)]$, to distinguish these
endpoints. Figure~\ref{fig:fraction_boundary} illustrates
the resulting boundary behavior. The original periodic
encoding produces relatively small sectors for targets
near $0^\circ$, but may generate narrow slices even when
a complete object ($360^\circ$) is requested.
The augmented encoding successfully generates complete
objects in these examples, but tends to produce
substantially oversized sectors for small target angles
($1^\circ$--$5^\circ$).

These qualitative results illustrate that resolving
the exact endpoint ambiguity does not necessarily
ensure robust control near both boundaries.
In particular, very small sectors remain challenging,
and our training data contains no examples below
$5^\circ$. We therefore retain the original periodic
encoding and leave improved boundary handling for
future work.

\subsection{Generation in Realistic Scenes}
\label{sec:supp_in_scene}

\begin{figure*}[t]
\centering
\newlength{\Sw}\setlength{\Sw}{3.10cm}   
\newcommand{\Simg}[1]{%
  \includegraphics[width=\Sw,height=\Sw,keepaspectratio=false]{figs/scene/#1}}
\newcommand{\Scell}[3]{
  \begin{minipage}[t]{\Sw}\centering
    \Simg{#1}\\[1.5pt]
    {\setlength{\baselineskip}{9pt}\scriptsize\color[gray]{0.25}#2\par}
    \vspace{-0.5pt}%
    {\scriptsize\bfseries\color[rgb]{0.10,0.10,0.60}#3}
  \end{minipage}}
\setlength{\tabcolsep}{2pt}
%
\subcaptionbox{%
  \textbf{Clock time control in realistic scenes.}
  Each image is generated from a unique scene prompt;
  the target time is shown in bold blue below each image.\label{fig:scene_clock}}{%
  \begin{tabular}{@{}ccccc@{}}
    \Scell{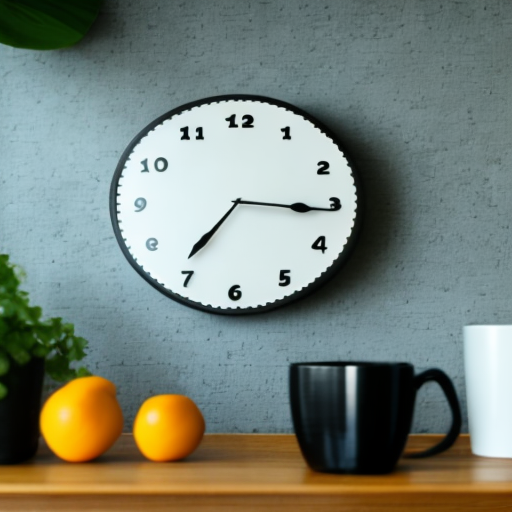}{in a bright kitchen with\\coffee mugs and fruit}{7:15} &
    \Scell{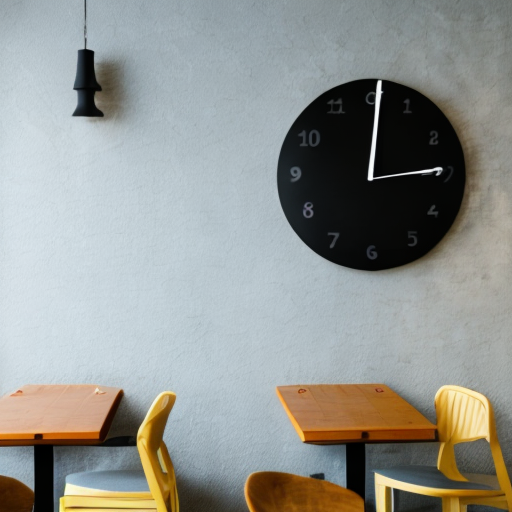}{on a cafe wall above\\small tables and chairs}{3:00} &
    \Scell{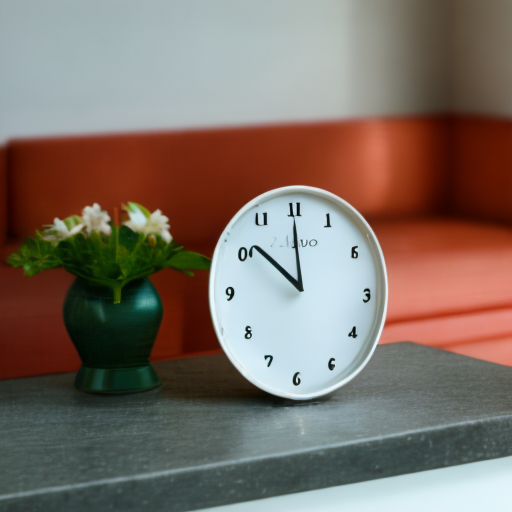}{in a small cafe with\\flowers on the counter}{10:00} &
    \Scell{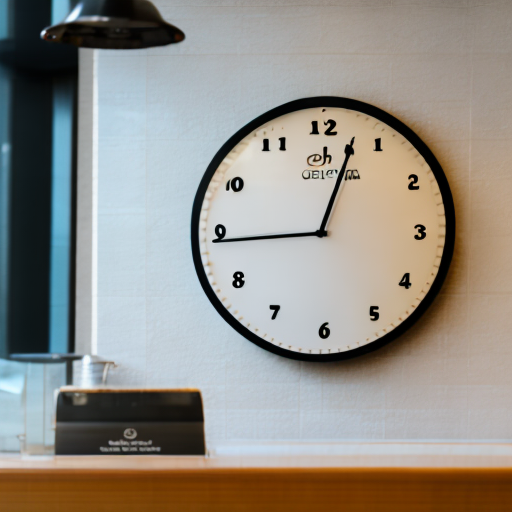}{behind a cafe counter with\\ coffee machine}{12:45} &
    \Scell{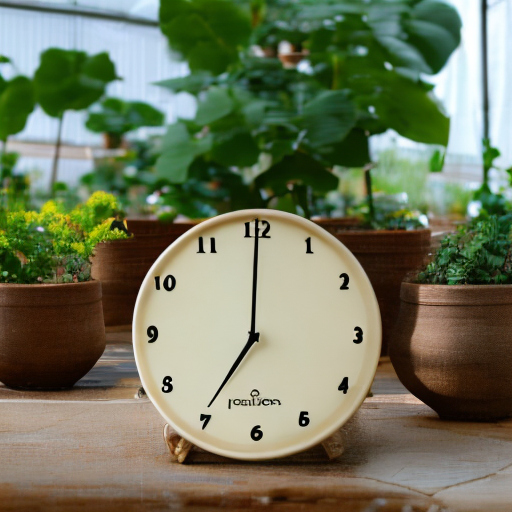}{in a greenhouse with\\plants and clay pots nearby}{7:00} \\\noalign{\vskip 6pt}
    \Scell{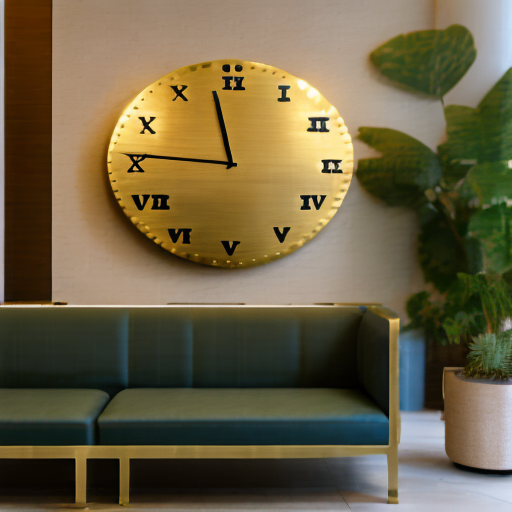}{in a hotel lobby with\\plants and a sofa nearby}{11:45} &
    \Scell{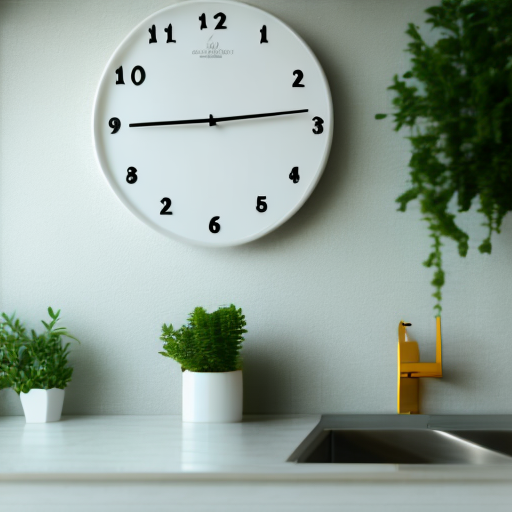}{in a modern kitchen with\\a sink and potted herbs}{2:45} &
    \Scell{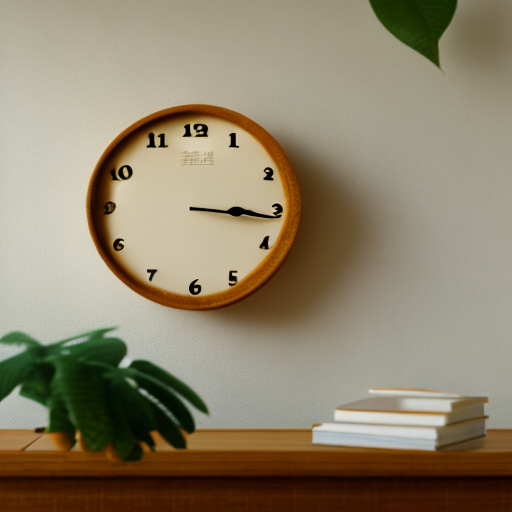}{in a living room with\\books and a leafy plant}{3:15} &
    \Scell{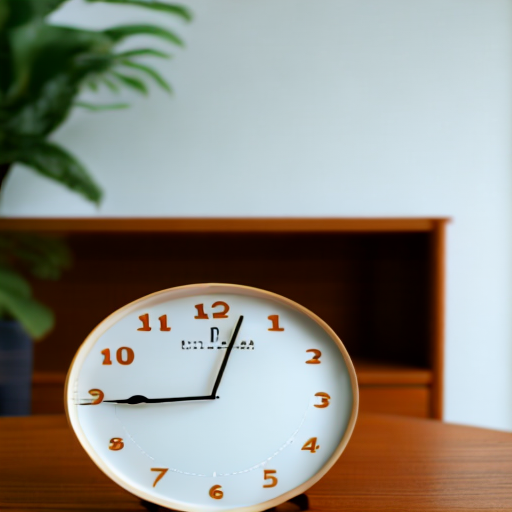}{near an office bookshelf\\with books and a plant}{12:45} &
    \Scell{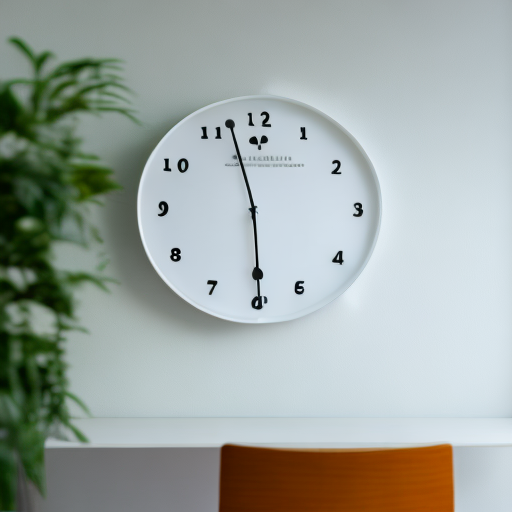}{on an office wall with\\a desk plant below}{11:30} \\
  \end{tabular}%
}
\\[8pt]
\subcaptionbox{%
  \textbf{Food-fraction control in top-down scenes.}
  The model renders the requested arc fraction across diverse food
  types and companion objects, with no bounding-box input at inference.\label{fig:scene_frac}}{%
  \begin{tabular}{@{}ccccc@{}}
    \Scell{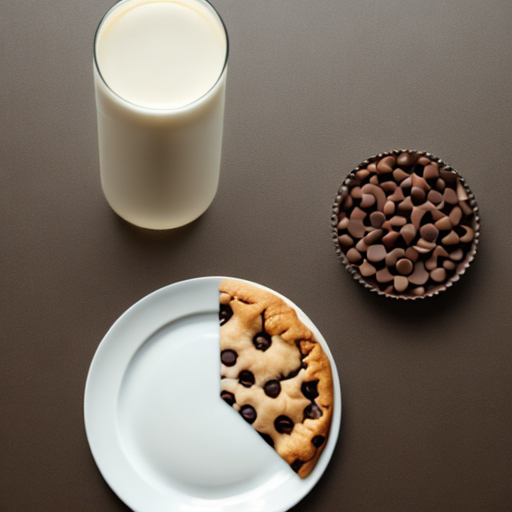}{giant chocolate chip cookie\\next to milk glass}{2/5} &
    \Scell{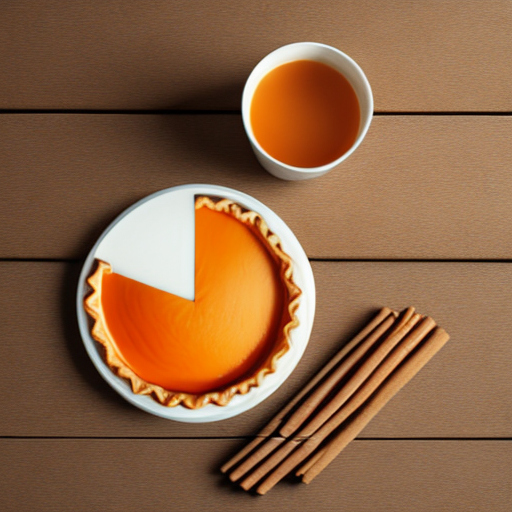}{round pumpkin pie next to\\cinnamon sticks and tea}{4/5} &
    \Scell{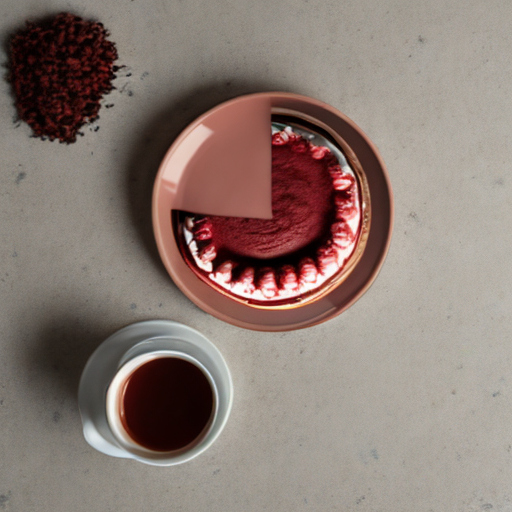}{round red velvet cake next\\to tea and berries}{3/4} &
    \Scell{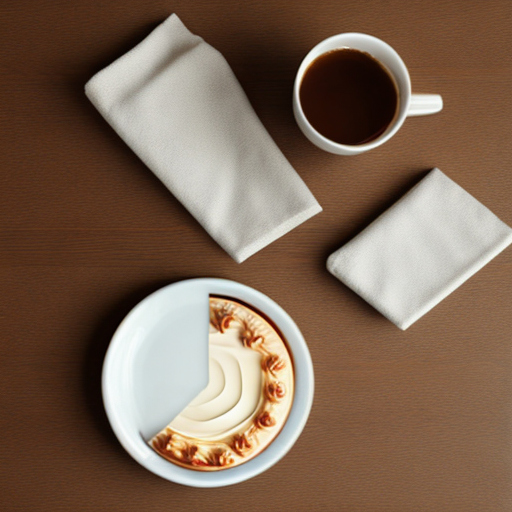}{round carrot cake next to\\coffee and folded napkin}{3/5} &
    \Scell{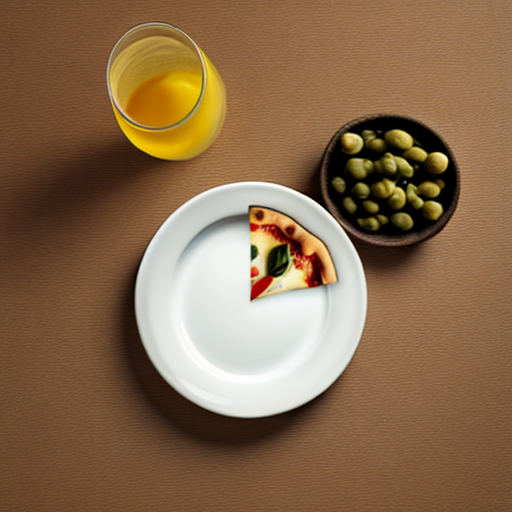}{round margherita pizza next\\to olive bowl and lemonade}{1/4} \\\noalign{\vskip 6pt}
    \Scell{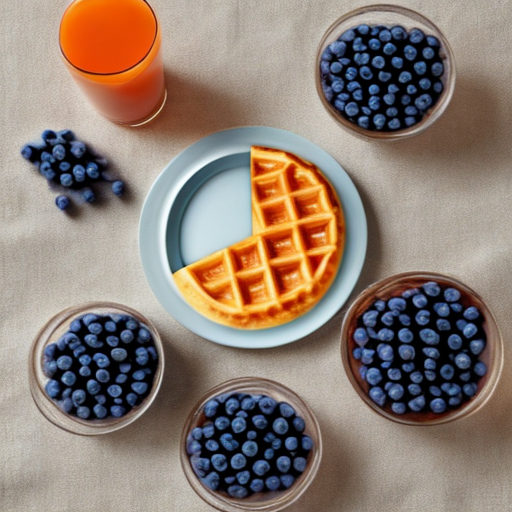}{round waffle next to\\blueberries and orange juice}{2/3} &
    \Scell{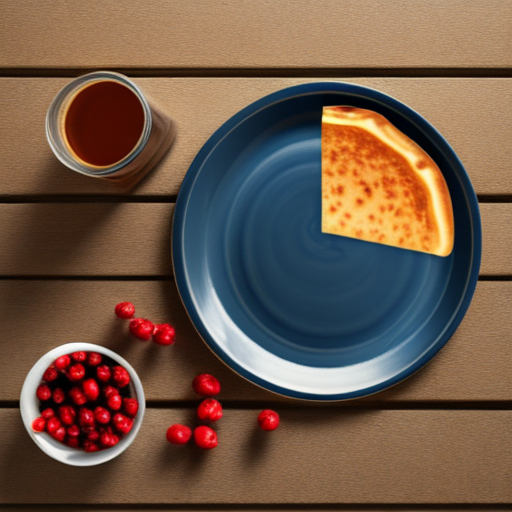}{round pancake next to\\syrup jar and berries}{1/4} &
    \Scell{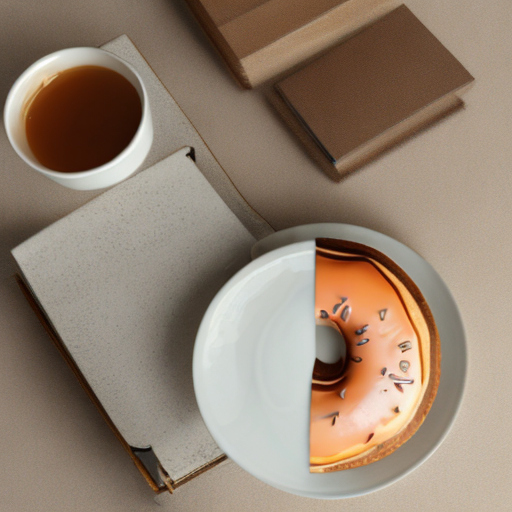}{powdered donut next to\\tea cup and open book}{1/2} &
    \Scell{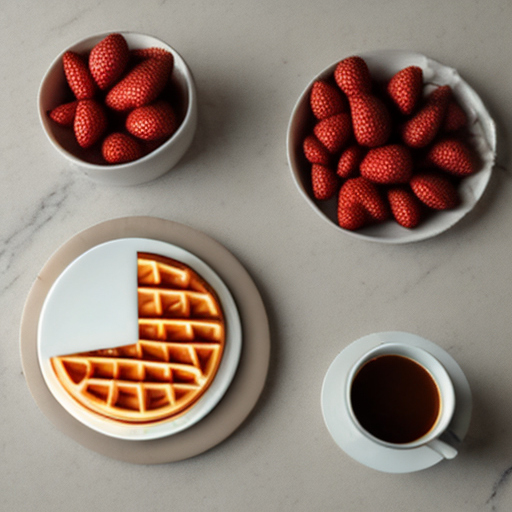}{Belgian waffle next to\\strawberries and espresso}{3/4} &
    \Scell{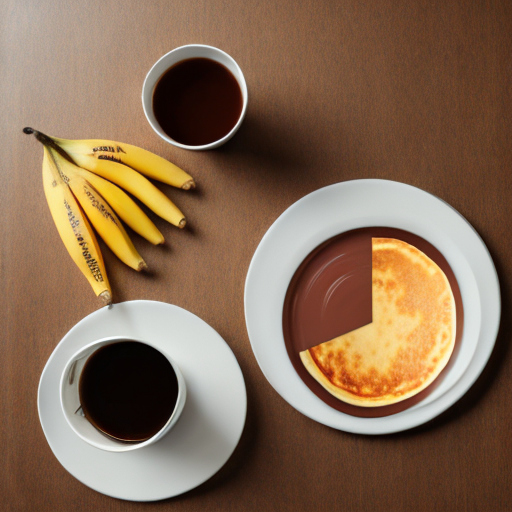}{large round pancake next\\to coffee and bananas}{2/3} \\
  \end{tabular}%
}
\caption{%
  \textbf{In-scene generation} with no bounding-box supervision at inference.
  \textbf{(a)} Clock time control across diverse room scenes.
  \textbf{(b)} Food-fraction control in top-down food settings.
  In both domains the model accurately tracks the requested
  geometric quantity (time / fraction) while maintaining
  scene coherence and object placement.
}
\label{fig:scene_generation}
\end{figure*}

We further evaluate both task-specific models in more complex scenes without
bounding-box input at inference. Figure~\ref{fig:scene_generation}(a) shows
clock-time control across diverse indoor environments, while
Fig.~\ref{fig:scene_generation}(b) shows fraction control for food objects in
scenes containing additional surrounding elements.

Despite changes in object scale, placement, and scene context, the models retain
the requested geometric control. Inference uses only the text prompt and target
quantity; no localization input or ground-truth bounding box is provided.

\end{document}